# A Framework for Sequential Planning in Multi-Agent Settings

**Piotr J. Gmytrasiewicz** PIOTR@CS.UIC.EDU
**Prashant Doshi** PDOSHI@CS.UIC.EDU
*Department of Computer Science*
*University of Illinois at Chicago*
*851 S. Morgan St*
*Chicago, IL 60607*

## Abstract

This paper extends the framework of partially observable Markov decision processes (POMDPs) to multi-agent settings by incorporating the notion of agent models into the state space. Agents maintain beliefs over physical states of the environment and over models of other agents, and they use Bayesian updates to maintain their beliefs over time. The solutions map belief states to actions. Models of other agents may include their belief states and are related to agent types considered in games of incomplete information. We express the agents' autonomy by postulating that their models are not directly manipulable or observable by other agents. We show that important properties of POMDPs, such as convergence of value iteration, the rate of convergence, and piece-wise linearity and convexity of the value functions carry over to our framework. Our approach complements a more traditional approach to interactive settings which uses Nash equilibria as a solution paradigm. We seek to avoid some of the drawbacks of equilibria which may be non-unique and do not capture off-equilibrium behaviors. We do so at the cost of having to represent, process and continuously revise models of other agents. Since the agent's beliefs may be arbitrarily nested, the optimal solutions to decision making problems are only asymptotically computable. However, approximate belief updates and approximately optimal plans are computable. We illustrate our framework using a simple application domain, and we show examples of belief updates and value functions.

## 1. Introduction

We develop a framework for sequential rationality of autonomous agents interacting with other agents within a common, and possibly uncertain, environment. We use the normative paradigm of decision-theoretic planning under uncertainty formalized as partially observable Markov decision processes (POMDPs) (Boutilier, Dean, & Hanks, 1999; Kaelbling, Littman, & Cassandra, 1998; Russell & Norvig, 2003) as a point of departure. Solutions of POMDPs are mappings from an agent's beliefs to actions. The drawback of POMDPs when it comes to environments populated by other agents is that other agents' actions have to be represented implicitly as environmental noise within the, usually static, transition model. Thus, an agent's beliefs about another agent are not part of solutions to POMDPs.

The main idea behind our formalism, called **interactive POMDPs** (I-POMDPs), is to allow agents to use more sophisticated constructs to model and predict behavior of other agents. Thus, we replace "flat" beliefs about the state space used in POMDPs with beliefs about the physical environment *and* about the other agent(s), possibly in terms of their preferences, capabilities, and beliefs. Such beliefs could include others' beliefs about others, and thus can be nested to arbitrary levels. They are called interactive beliefs. While the space of interactive beliefs is very rich and updating these beliefs is more complex than updating their "flat" counterparts, we use the value function plots to show that solutions to I-POMDPs are at least as good as, and in usual cases superior to, comparable solutions to POMDPs. The reason is intuitive – maintaining sophisticated models of other agents allows more refined analysis of their behavior and better predictions of their actions.

I-POMDPs are applicable to autonomous self-interested agents who locally compute what actions they should execute to optimize their preferences given what they believe while interacting with others with possibly conflicting objectives. Our approach of using a decision-theoretic framework and solution concept complements the equilibrium approach to analyzing interactions as used in classical game theory (Fudenberg & Tirole, 1991). The drawback of equilibria is that there could be many of them (non-uniqueness), and that they describe agent's optimal actions only if, and when, an equilibrium has been reached (incompleteness). Our approach, instead, is centered on optimality and best response to anticipated action of other agent(s), rather then on stability (Binmore, 1990; Kadane & Larkey, 1982). The question of whether, under what circumstances, and what kind of equilibria could arise from solutions to I-POMDPs is currently open.

Our approach avoids the difficulties of non-uniqueness and incompleteness of traditional equilibrium approach, and offers solutions which are likely to be better than the solutions of traditional POMDPs applied to multi-agent settings. But these advantages come at the cost of processing and maintaining possibly infinitely nested interactive beliefs. Consequently, only approximate belief updates and approximately optimal solutions to planning problems are computable in general. We define a class of finitely nested I-POMDPs to form a basis for computable approximations to infinitely nested ones. We show that a number of properties that facilitate solutions of POMDPs carry over to finitely nested I-POMDPs. In particular, the interactive beliefs are sufficient statistics for the histories of agent's observations, the belief update is a generalization of the update in POMDPs, the value function is piece-wise linear and convex, and the value iteration algorithm converges at the same rate.

The remainder of this paper is structured as follows. We start with a brief review of related work in Section 2, followed by an overview of partially observable Markov decision processes in Section 3. There, we include a simple example of a tiger game. We introduce the concept of agent types in Section 4. Section 5 introduces interactive POMDPs and defines their solutions. The finitely nested I-POMDPs, and some of their properties are introduced in Section 6. We continue with an example application of finitely nested I-POMDPs to a multi-agent version of the tiger game in Section 7. There, we show examples of belief updates and value functions. We conclude with a brief summary and some current research issues in Section 8. Details of all proofs are in the Appendix.

## 2. Related Work

Our work draws from prior research on partially observable Markov decision processes, which recently gained a lot of attention within the AI community (Smallwood & Sondik, 1973; Monahan, 1982; Lovejoy, 1991; Hausktecht, 1997; Kaelbling et al., 1998; Boutilier et al., 1999; Hauskrecht, 2000).

The formalism of Markov decision processes has been extended to multiple agents giving rise to stochastic games or Markov games (Fudenberg & Tirole, 1991). Traditionally, the solution concept used for stochastic games is that of Nash equilibria. Some recent work in AI follows that tradition (Littman, 1994; Hu & Wellman, 1998; Boutilier, 1999; Koller & Milch, 2001). However, as we mentioned before, and as has been pointed out by some game theorists (Binmore, 1990; Kadane &

Larkey, 1982), while Nash equilibria are useful for describing a multi-agent system when, and if, it has reached a stable state, this solution concept is not sufficient as a general control paradigm. The main reasons are that there may be multiple equilibria with no clear way to choose among them (non-uniqueness), and the fact that equilibria do not specify actions in cases in which agents believe that other agents may not act according to their equilibrium strategies (incompleteness).

Other extensions of POMDPs to multiple agents appeared in AI literature recently (Bernstein, Givan, Immerman, & Zilberstein, 2002; Nair, Pynadath, Yokoo, Tambe, & Marsella, 2003). They have been called decentralized POMDPs (DEC-POMDPs), and are related to decentralized control problems (Ooi & Wornell, 1996). DEC-POMDP framework assumes that the agents are fully cooperative, i.e., they have common reward function and form a team. Furthermore, it is assumed that the optimal joint solution is computed centrally and then distributed among the agents for execution.

From the game-theoretic side, we are motivated by the subjective approach to probability in games (Kadane & Larkey, 1982), Bayesian games of incomplete information (see Fudenberg & Tirole, 1991; Harsanyi, 1967, and references therein), work on interactive belief systems (Harsanyi, 1967; Mertens & Zamir, 1985; Brandenburger & Dekel, 1993; Fagin, Halpern, Moses, & Vardi, 1995; Aumann, 1999; Fagin, Geanakoplos, Halpern, & Vardi, 1999), and insights from research on learning in game theory (Fudenberg & Levine, 1998). Our approach, closely related to decision-theoretic (Myerson, 1991), or epistemic (Ambruster & Boge, 1979; Battigalli & Siniscalchi, 1999; Brandenburger, 2002) approach to game theory, consists of predicting actions of other agents given all available information, and then of choosing the agent's own action (Kadane & Larkey, 1982). Thus, the descriptive aspect of decision theory is used to predict others' actions, and its prescriptive aspect is used to select agent's own optimal action.

The work presented here also extends previous work on Recursive Modeling Method (RMM) (Gmytrasiewicz & Durfee, 2000), but adds elements of belief update and sequential planning.

## 3. Background: Partially Observable Markov Decision Processes

A partially observable Markov decision process (POMDP) (Monahan, 1982; Hausktecht, 1997; Kaelbling et al., 1998; Boutilier et al., 1999; Hauskrecht, 2000) of an agent $i$ is defined as

$$\mathit{POMDP}_i = \langle S, A_i, T_i, \Omega_i, O_i, R_i \rangle \tag{1}$$

where: $S$ is a set of possible states of the environment. $A_i$ is a set of actions agent $i$ can execute. $T_i$ is a transition function – $T_i : S \times A_i \times S \rightarrow [0, 1]$ which describes results of agent $i$'s actions. $\Omega_i$ is the set of observations the agent $i$ can make. $O_i$ is the agent's observation function – $O_i : S \times A_i \times \Omega_i \rightarrow [0, 1]$ which specifies probabilities of observations given agent's actions and resulting states. Finally, $R_i$ is the reward function representing the agent $i$'s preferences – $R_i : S \times A_i \rightarrow \Re$.

In POMDPs, an agent's belief about the state is represented as a probability distribution over $S$. Initially, before any observations or actions take place, the agent has some (prior) belief, $b_i^0$. After some time steps, $t$, we assume that the agent has $t + 1$ observations and has performed $t$ actions[1]. These can be assembled into *agent $i$'s observation history*: $h_i^t = \{o_i^0, o_i^1, .., o_i^{t-1}, o_i^t\}$ at time $t$. Let $H_i$ denote the set of all observation histories of agent $i$. The agent's current belief, $b_i^t$ over $S$, is continuously revised based on new observations and expected results of performed actions. It turns

1. We assume that action is taken at every time step; it is without loss of generality since any of the actions maybe a No-op.

out that the agent's belief state is sufficient to summarize all of the past observation history and initial belief; hence it is called a sufficient statistic.[2]

The belief update takes into account changes in initial belief, $b_i^{t-1}$, due to action, $a_i^{t-1}$, executed at time $t-1$, and the new observation, $o_i^t$. The new belief, $b_i^t$, that the current state is $s^t$, is:

$$b_i^t(s^t) = \beta O_i(o_i^t, s^t, a_i^{t-1}) \sum_{s^{t-1} \in S} b_i^{t-1}(s^{t-1}) T_i(s^t, a_i^t, s^{t-1}) \tag{2}$$

where $\beta$ is the normalizing constant.

It is convenient to summarize the above update performed for all states in $S$ as $b_i^t = SE(b_i^{t-1}, a_i^{t-1}, o_i^t)$ (Kaelbling et al., 1998).

### 3.1 Optimality Criteria and Solutions

The agent's optimality criterion, $OC_i$, is needed to specify how rewards acquired over time are handled. Commonly used criteria include:

- A finite horizon criterion, in which the agent maximizes the expected value of the sum of the following $T$ rewards: $E(\sum_{t=0}^{T} r_t)$. Here, $r_t$ is a reward obtained at time $t$ and $T$ is the length of the horizon. We will denote this criterion as *fh*$^T$.
- An infinite horizon criterion with discounting, according to which the agent maximizes $E(\sum_{t=0}^{\infty} \gamma^t r_t)$, where $0 < \gamma < 1$ is a discount factor. We will denote this criterion as *ih*$^\gamma$.
- An infinite horizon criterion with averaging, according to which the agent maximizes the average reward per time step. We will denote this as *ih*$^{AV}$.

In what follows, we concentrate on the infinite horizon criterion with discounting, but our approach can be easily adapted to the other criteria.

The utility associated with a belief state, $b_i$ is composed of the best of the immediate rewards that can be obtained in $b_i$, together with the discounted expected sum of utilities associated with belief states following $b_i$:

$$U(b_i) = \max_{a_i \in A_i} \left\{ \sum_{s \in S} b_i(s) R_i(s, a_i) + \gamma \sum_{o_i \in \Omega_i} Pr(o_i | a_i, b_i) U(SE_i(b_i, a_i, o_i)) \right\} \tag{3}$$

Value iteration uses the Equation 3 iteratively to obtain values of belief states for longer time horizons. At each step of the value iteration the error of the current value estimate is reduced by the factor of at least $\gamma$ (see for example Russell & Norvig, 2003, Section 17.2.) The optimal action, $a_i^*$, is then an element of the set of optimal actions, $OPT(b_i)$, for the belief state, defined as:

$$OPT(b_i) = \underset{a_i \in A_i}{argmax} \left\{ \sum_{s \in S} b_i(s) R_i(s, a_i) + \gamma \sum_{o_i \in \Omega_i} Pr(o_i | a_i, b_i) U(SE(b_i, a_i, o_i)) \right\} \tag{4}$$

2. See (Smallwood & Sondik, 1973) for proof.

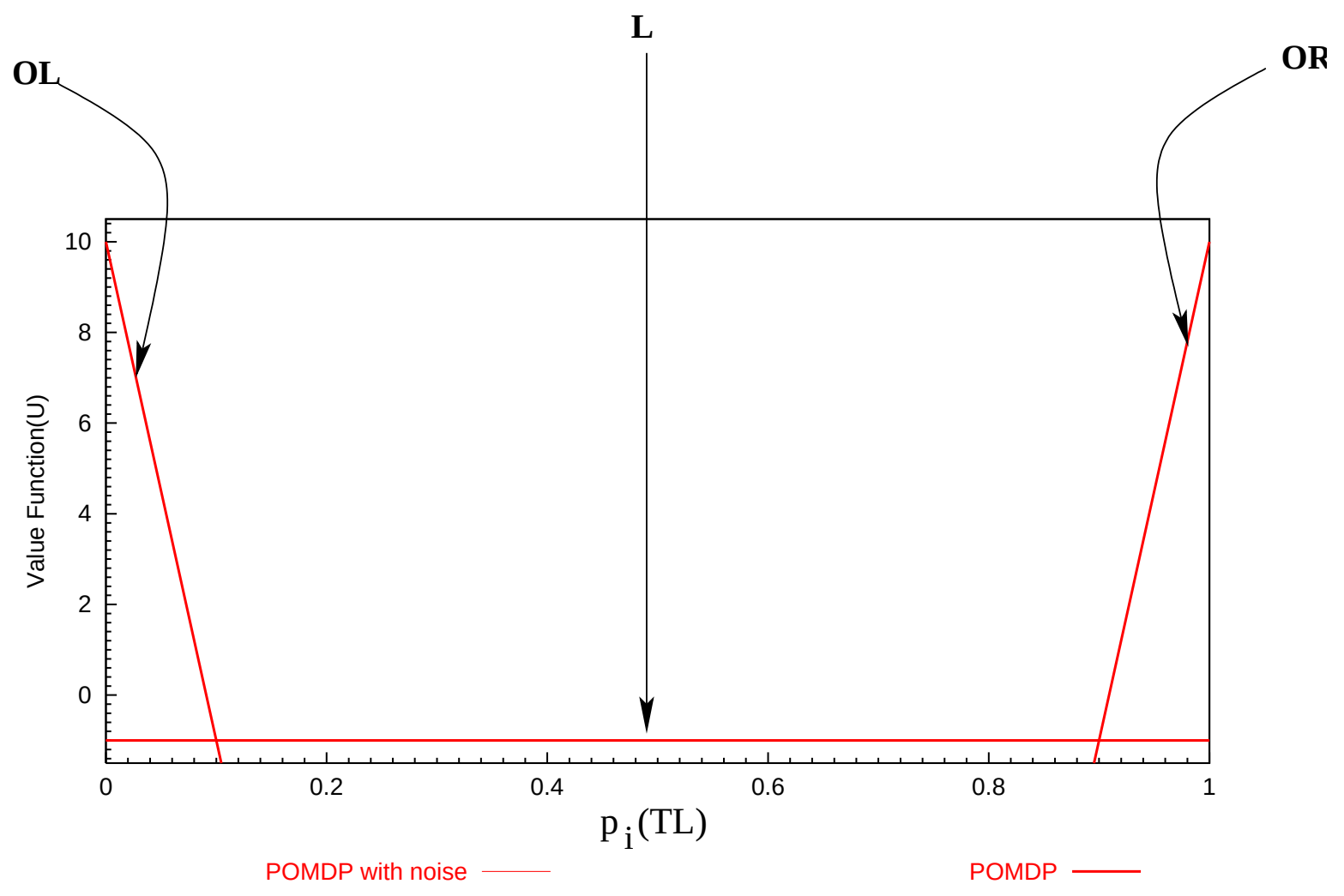

Figure 1: The value function for single agent tiger game with time horizon of length 1, $OC_i = fh^1$. Actions are: open right door - OR, open left door - OL, and listen - L. For this value of the time horizon the value function for a POMDP with noise factor is identical to single agent POMDP.

### 3.2 Example: The Tiger Game

We briefly review the POMDP solutions to the tiger game (Kaelbling et al., 1998). Our purpose is to build on the insights that POMDP solutions provide in this simple case to illustrate solutions to interactive versions of this game later.

The traditional tiger game resembles a game-show situation in which the decision maker has to choose to open one of two doors behind which lies either a valuable prize or a dangerous tiger. Apart from actions that open doors, the subject has the option of listening for the tiger's growl coming from the left, or the right, door. However, the subject's hearing is imperfect, with given percentages (say, 15%) of false positive and false negative occurrences. Following (Kaelbling et al., 1998), we assume that the value of the prize is 10, that the pain associated with encountering the tiger can be quantified as -100, and that the cost of listening is -1.

The value function, in Figure 1, shows values of various belief states when the agent's time horizon is equal to 1. Values of beliefs are based on best action available in that belief state, as specified in Eq. 3. The state of certainty is most valuable – when the agent knows the location of the tiger it can open the opposite door and claim the prize which certainly awaits. Thus, when the probability of tiger location is 0 or 1, the value is 10. When the agent is sufficiently uncertain, its best option is to play it safe and listen; the value is then -1. The agent is indifferent between opening doors and listening when it assigns probabilities of 0.9 or 0.1 to the location of the tiger.

Note that, when the time horizon is equal to 1, listening does not provide any useful information since the game does not continue to allow for the use of this information. For longer time horizons the benefits of results of listening results in policies which are better in some ranges of initial belief. Since the value function is composed of values corresponding to actions, which are linear in prob-ability of tiger location, the value function has the property of being piece-wise linear and convex (PWLC) for all horizons. This simplifies the computations substantially.

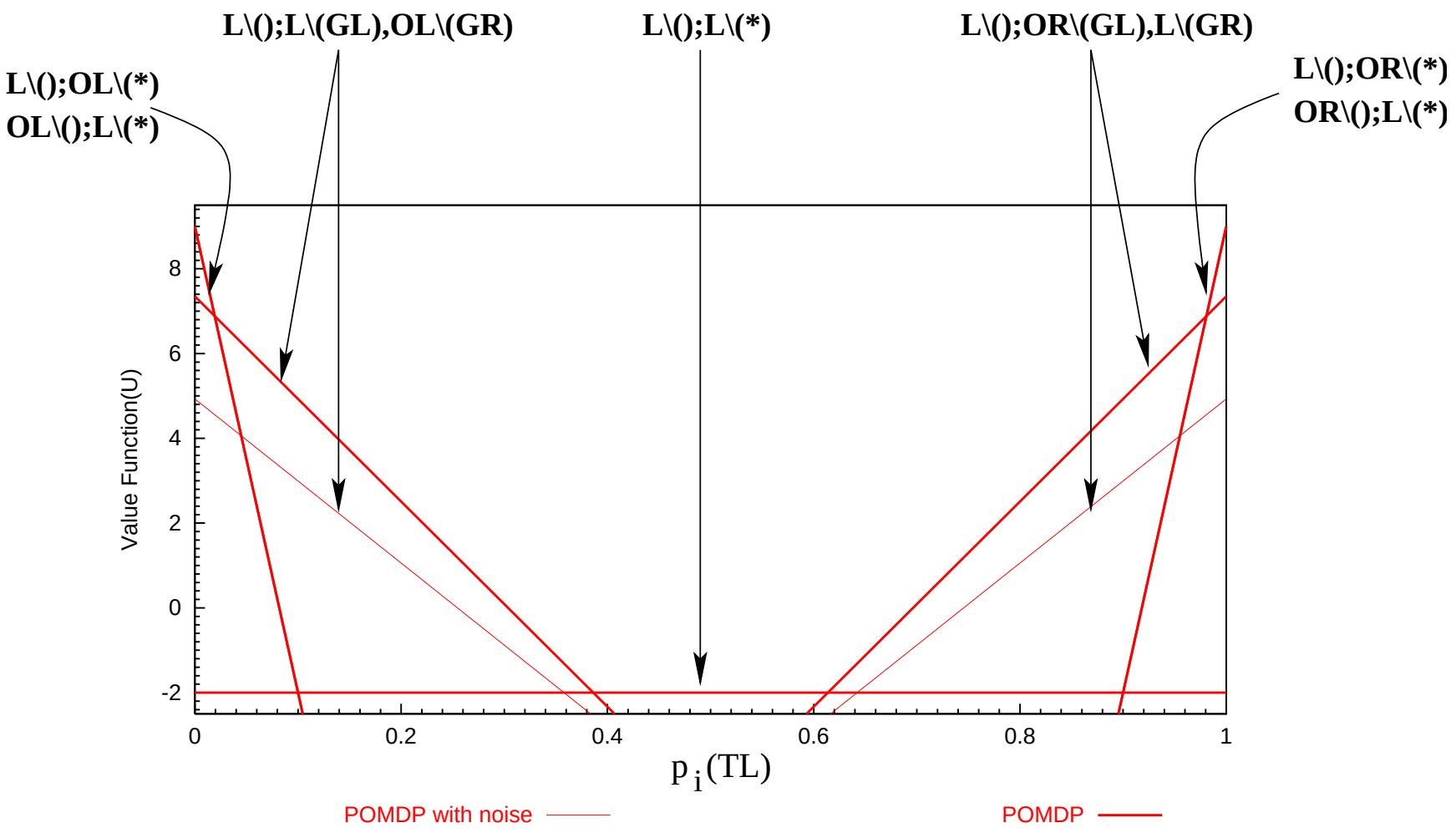

Figure 2: The value function for single agent tiger game compared to an agent facing a noise factor, for horizon of length 2. Policies corresponding to value lines are conditional plans. Actions, L, OR or OL, are conditioned on observational sequences in parenthesis. For example L\();L\(GL),OL\(GR) denotes a plan to perform the listening action, L, at the beginning (list of observations is empty), and then another L if the observation is growl from the left (GL), and open the left door, OL, if the observation is GR. $*$ is a wildcard with the usual interpretation.

In Figure 2 we present a comparison of value functions for horizon of length 2 for a single agent, and for an agent facing a more noisy environment. The presence of such noise could be due to another agent opening the doors or listening with some probabilities.[3] Since POMDPs do not include explicit models of other agents, these noise actions have been included in the transition model, $T$.

Consequences of folding noise into $T$ are two-fold. First, the effectiveness of the agent's optimal policies declines since the value of hearing growls diminishes over many time steps. Figure 3 depicts a comparison of value functions for horizon of length 3. Here, for example, two consecutive growls in a noisy environment are not as valuable as when the agent knows it is acting alone since the noise may have perturbed the state of the system between the growls. For time horizon of length 1 the noise does not matter and the value vectors overlap, as in Figure 1.

Second, since the presence of another agent is implicit in the static transition model, the agent cannot update its model of the other agent's actions during repeated interactions. This effect becomes more important as time horizon increases. Our approach addresses this issue by allowing explicit modeling of the other agent(s). This results in policies of superior quality, as we show in Section 7. Figure 4 shows a policy for an agent facing a noisy environment for time horizon of 3. We compare it to the corresponding I-POMDP policy in Section 7. Note that it is slightly different

3. We assumed that, due to the noise, either door opens with probabilities of 0.1 at each turn, and nothing happens with the probability 0.8. We explain the origin of this assumption in Section 7.

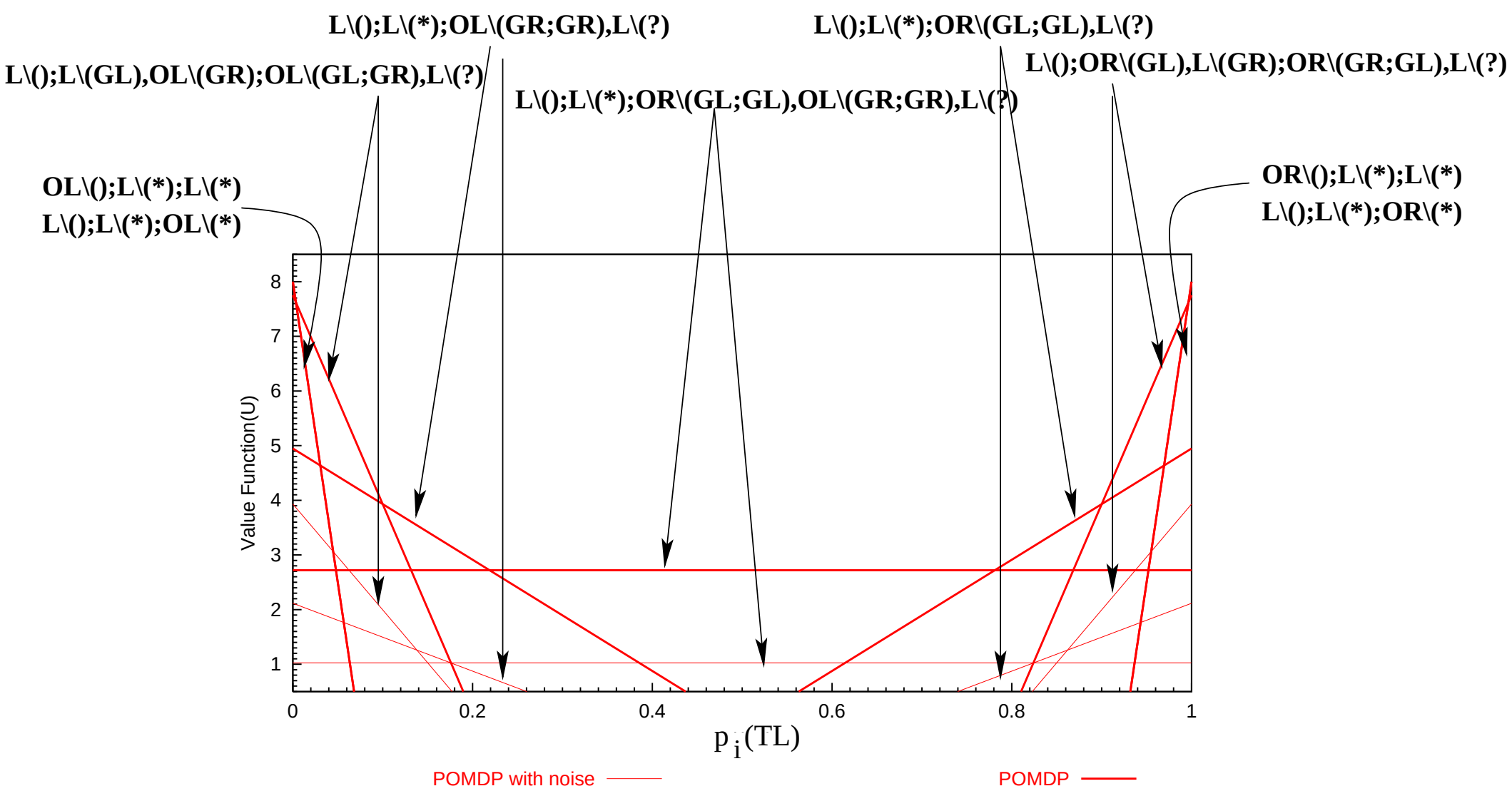

Figure 3: The value function for single agent tiger game compared to an agent facing a noise factor, for horizon of length 3. The “?” in the description of a policy stands for any of the perceptual sequences not yet listed in the description of the policy.

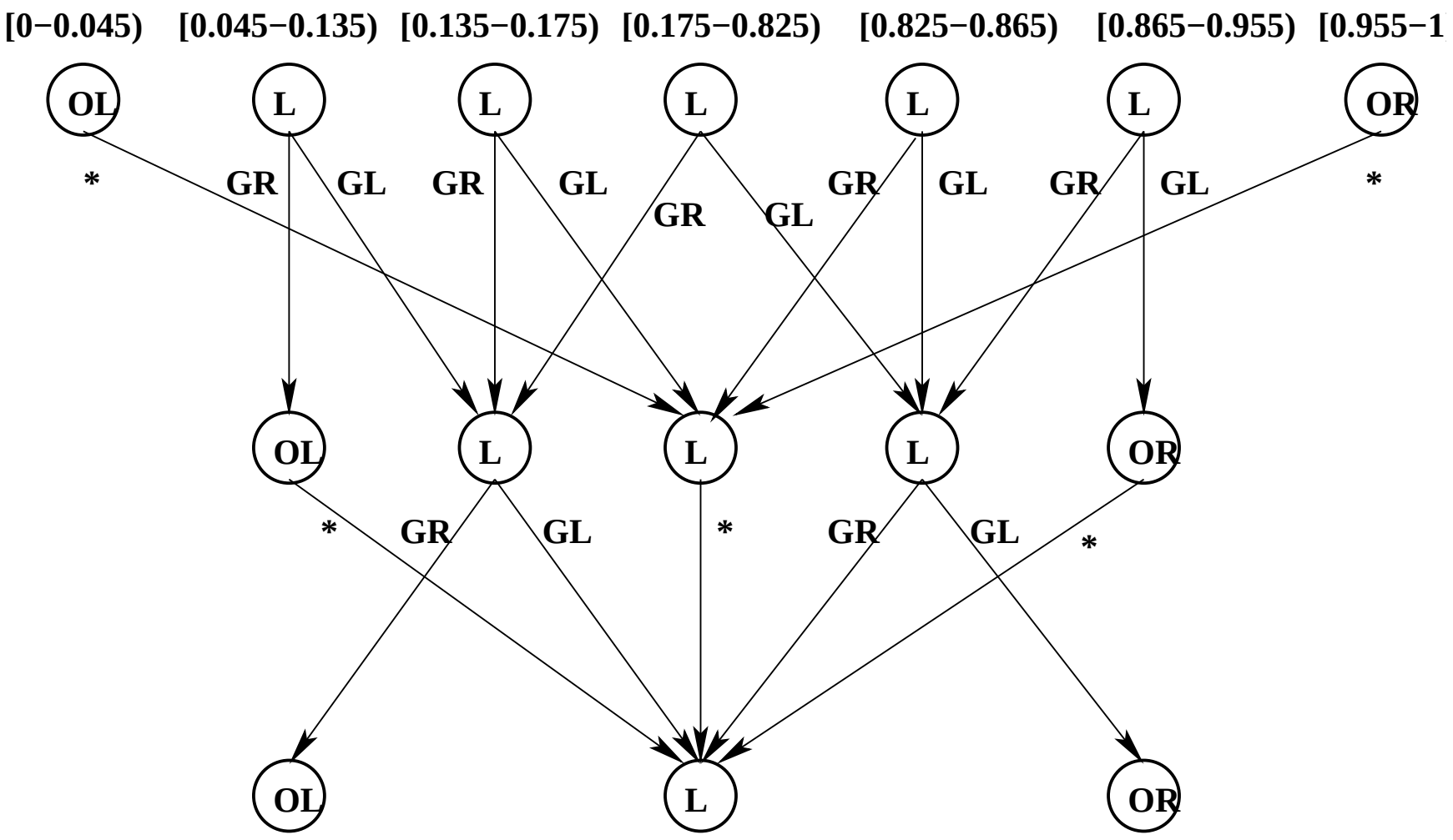

Figure 4: The policy graph corresponding to value function of POMDP with noise depicted in Fig. 3.

than the policy without noise in the example by Kaelbling, Littman and Cassandra (1998) due to differences in value functions.

## 4. Agent Types and Frames

The POMDP definition includes parameters that permit us to compute an agent's optimal behavior,[4] conditioned on its beliefs. Let us collect these implementation independent factors into a construct we call an agent $i$'s *type*.

**Definition 1 (Type).** *A type of an agent $i$ is, $\theta_i = \langle b_i, A_i, \Omega_i, T_i, O_i, R_i, OC_i \rangle$, where $b_i$ is agent $i$'s state of belief (an element of $\Delta(S)$), $OC_i$ is its optimality criterion, and the rest of the elements are as defined before. Let $\Theta_i$ be the set of agent $i$'s types.*

Given type, $\theta_i$, and the assumption that the agent is Bayesian-rational, the set of agent's optimal actions will be denoted as $OPT(\theta_i)$. In the next section, we generalize the notion of type to situations which include interactions with other agents; it then coincides with the notion of type used in Bayesian games (Fudenberg & Tirole, 1991; Harsanyi, 1967).

It is convenient to define the notion of a *frame*, $\widehat{\theta}_i$, of agent $i$:

**Definition 2 (Frame).** *A frame of an agent $i$ is, $\widehat{\theta}_i = \langle A_i, \Omega_i, T_i, O_i, R_i, OC_i \rangle$. Let $\widehat{\Theta}_i$ be the set of agent $i$'s frames.*

For brevity one can write a type as consisting of an agent's belief together with its frame: $\theta_i = \langle b_i, \widehat{\theta}_i \rangle$.

In the context of the tiger game described in the previous section, agent type describes the agent's actions and their results, the quality of the agent's hearing, its payoffs, and its belief about the tiger location.

Realistically, apart from implementation-independent factors grouped in type, an agent's behavior may also depend on implementation-specific parameters, like the processor speed, memory available, etc. These can be included in the (implementation dependent, or *complete*) type, increasing the accuracy of predicted behavior, but at the cost of additional complexity. Definition and use of complete types is a topic of ongoing work.

## 5. Interactive POMDPs

As we mentioned, our intention is to generalize POMDPs to handle presence of other agents. We do this by including descriptions of other agents (their types for example) in the state space. For simplicity of presentation, we consider an agent $i$, that is interacting with one other agent, $j$. The formalism easily generalizes to larger number of agents.

**Definition 3 (I-POMDP).** *An* interactive POMDP *of agent $i$, I-POMDP$_i$, is:*

$$\text{I-POMDP}_i = \langle IS_i, A, T_i, \Omega_i, O_i, R_i \rangle \tag{5}$$

4. The issue of computability of solutions to POMDPs has been a subject of much research (Papadimitriou & Tsitsiklis, 1987; Madani, Hanks, & Condon, 2003). It is of obvious importance when one uses POMDPs to model agents; we return to this issue later.

where:

• $IS_i$ is a set of **interactive** states defined as $IS_i = S \times M_j$,[5] interacting with agent $i$, where $S$ is the set of states of the physical environment, and $M_j$ is the set of possible models of agent $j$. Each model, $m_j \in M_j$, is defined as a triple $m_j = \langle h_j, f_j, O_j \rangle$, where $f_j : H_j \to \Delta(A_j)$ is agent $j$'s function, assumed computable, which maps possible histories of $j$'s observations to distributions over its actions. $h_j$ is an element of $H_j$, and $O_j$ is a function specifying the way the environment is supplying the agent with its input. Sometimes we write model $m_j$ as $m_j = \langle h_j, \widehat{m}_j \rangle$, where $\widehat{m}_j$ consists of $f_j$ and $O_j$. It is convenient to subdivide the set of models into two classes. The subintentional models, $SM_j$, are relatively simple, while the intentional models, $IM_j$, use the notion of rationality to model the other agent. Thus, $M_j = IM_j \cup SM_j$.

Simple examples of subintentional models include a no-information model and a fictitious play model, both of which are history independent. A no-information model (Gmytrasiewicz & Durfee, 2000) assumes that each of the other agent's actions is executed with equal probability. Fictitious play (Fudenberg & Levine, 1998) assumes that the other agent chooses actions according to a fixed but unknown distribution, and that the original agent's prior belief over that distribution takes a form of a Dirichlet distribution.[6] An example of a more powerful subintentional model is a finite state controller.

The intentional models are more sophisticated in that they ascribe to the other agent beliefs, preferences and rationality in action selection.[7] Intentional models are thus $j$'s types, $\theta_j = \langle b_j, \widehat{\theta}_j \rangle$, under the assumption that agent $j$ is Bayesian-rational.[8] Agent $j$'s belief is a probability distribution over states of the environment and the models of the agent $i$; $b_j \in \Delta(S \times M_i)$. The notion of a type we use here coincides with the notion of type in game theory, where it is defined as consisting of all of the agent $i$'s private information relevant to its decision making (Harsanyi, 1967; Fudenberg & Tirole, 1991). In particular, if agents' beliefs are private information, then their types involve possibly infinitely nested beliefs over others' types and their beliefs about others (Mertens & Zamir, 1985; Brandenburger & Dekel, 1993; Aumann, 1999; Aumann & Heifetz, 2002).[9] They are related to recursive model structures in our prior work (Gmytrasiewicz & Durfee, 2000). The definition of interactive state space is consistent with the notion of a completely specified state space put forward by Aumann (1999). Similar state spaces have been proposed by others (Mertens & Zamir, 1985; Brandenburger & Dekel, 1993).

• $A = A_i \times A_j$ is the set of joint moves of all agents.

• $T_i$ is the transition model. The usual way to define the transition probabilities in POMDPs is to assume that the agent's actions can change any aspect of the state description. In case of I-POMDPs, this would mean actions modifying any aspect of the interactive states, including other agents' observation histories and their functions, or, if they are modeled intentionally, their beliefs and reward functions. Allowing agents to directly manipulate other agents in such ways, however, violates the notion of agents' autonomy. Thus, we make the following simplifying assumption:

5. If there are more agents, say $N > 2$, then $IS_i = S \times_{j=1}^{N-1} M_j$
6. Technically, according to our notation, fictitious play is actually an ensemble of models.
7. Dennet (1986) advocates ascribing rationality to other agent(s), and calls it ”assuming an intentional stance towards them”.
8. Note that the space of types is by far richer than that of computable models. In particular, since the set of computable models is countable and the set of types is uncountable, many types are not computable models.
9. Implicit in the definition of interactive beliefs is the assumption of coherency (Brandenburger & Dekel, 1993).

**Model Non-manipulability Assumption (MNM):** Agents' actions do not change the other agents' models directly.

Given this simplification, the transition model can be defined as $T_i : S \times A \times S \rightarrow [0, 1]$

Autonomy, formalized by the MNM assumption, precludes, for example, direct "mind control", and implies that other agents' belief states can be changed only indirectly, typically by changing the environment in a way observable to them. In other words, agents' beliefs change, like in POMDPs, but as a result of belief update after an observation, not as a direct result of any of the agents' actions.[10]

• $\Omega_i$ is defined as before in the POMDP model.

• $O_i$ is an observation function. In defining this function we make the following assumption:

**Model Non-observability (MNO):** Agents cannot observe other's models directly.

Given this assumption the observation function is defined as $O_i : S \times A \times \Omega_i \rightarrow [0, 1]$.

The MNO assumption formalizes another aspect of autonomy – agents are autonomous in that their observations and functions, or beliefs and other properties, say preferences, in intentional models, are private and the other agents cannot observe them directly.[11]

• $R_i$ is defined as $R_i : IS_i \times A \rightarrow \Re$. We allow the agent to have preferences over physical states and models of other agents, but usually only the physical state will matter.

As we mentioned, we see interactive POMDPs as a subjective counterpart to an objective external view in stochastic games (Fudenberg & Tirole, 1991), and also followed in some work in AI (Boutilier, 1999) and (Koller & Milch, 2001) and in decentralized POMDPs (Bernstein et al., 2002; Nair et al., 2003). Interactive POMDPs represent an individual agent's point of view on the environment and the other agents, and facilitate planning and problem solving at the agent's own individual level.

### 5.1 Belief Update in *I-POMDPs*

We will show that, as in POMDPs, an agent's beliefs over their interactive states are sufficient statistics, i.e., they fully summarize the agent's observation histories. Further, we need to show how beliefs are updated after the agent's action and observation, and how solutions are defined.

The new belief state, $b_i^t$, is a function of the previous belief state, $b_i^{t-1}$, the last action, $a_i^{t-1}$, and the new observation, $o_i^t$, just as in POMDPs. There are two differences that complicate belief update when compared to POMDPs. First, since the state of the physical environment depends on the actions performed by both agents the prediction of how the physical state changes has to be made based on the probabilities of various actions of the other agent. The probabilities of other's actions are obtained based on their models. Thus, unlike in Bayesian and stochastic games, we do not assume that actions are fully observable by other agents. Rather, agents can attempt to infer what actions other agents have performed by sensing their results on the environment. Second, changes in the models of other agents have to be included in the update. These reflect the other's observations and, if they are modeled intentionally, the update of the other agent's beliefs. In this case, the agent has to update its beliefs about the other agent based on what it anticipates the other agent observes

10. The possibility that agents can influence the observational capabilities of other agents can be accommodated by including the factors that can change sensing capabilities in the set $S$.

11. Again, the possibility that agents can observe factors that may influence the observational capabilities of other agents is allowed by including these factors in $S$.

and how it updates. As could be expected, the update of the possibly infinitely nested belief over other's types is, in general, only asymptotically computable.

**Proposition 1. (Sufficiency)** *In an interactive POMDP of agent $i$, $i$'s current belief, i.e., the probability distribution over the set $S \times M_j$, is a sufficient statistic for the past history of $i$'s observations.*

The next proposition defines the agent $i$'s belief update function, $b_i^t(is^t) = Pr(is^t|o_i^t, a_i^{t-1}, b_i^{t-1})$, where $is^t \in IS_i$ is an interactive state. We use the belief state estimation function, $SE_{\theta_i}$, as an abbreviation for belief updates for individual states so that $b_i^t = SE_{\theta_i}(b_i^{t-1}, a_i^{t-1}, o_i^t)$. $\tau_{\theta_i}(b_i^{t-1}, a_i^{t-1}, o_i^t, b_i^t)$ will stand for $Pr(b_i^t|b_i^{t-1}, a_i^{t-1}, o_i^t)$. Further below we also define the set of type-dependent optimal actions of an agent, $OPT(\theta_i)$.

**Proposition 2. (Belief Update)** *Under the MNM and MNO assumptions, the belief update function for an interactive POMDP $\langle IS_i, A, T_i, \Omega_i, O_i, R_i \rangle$, when $m_j$ in $is^t$ is intentional, is:*

$$\begin{aligned} b_i^t(is^t) &= \beta \sum_{is^{t-1}:\widehat{m}_j^{t-1}=\widehat{\theta}_j^t} b_i^{t-1}(is^{t-1}) \sum_{a_j^{t-1}} Pr(a_j^{t-1}|\theta_j^{t-1}) O_i(s^t, a^{t-1}, o_i^t) \\ &\quad \times T_i(s^{t-1}, a^{t-1}, s^t) \sum_{o_j^t} \tau_{\theta_j^t}(b_j^{t-1}, a_j^{t-1}, o_j^t, b_j^t) O_j(s^t, a^{t-1}, o_j^t) \end{aligned} \tag{6}$$

*When $m_j$ in $is^t$ is subintentional the first summation extends over $is^{t-1} : \widehat{m}_j^{t-1} = \widehat{m}_j^t$, $Pr(a_j^{t-1}|\theta_j^{t-1})$ is replaced with $Pr(a_j^{t-1}|m_j^{t-1})$, and $\tau_{\theta_j^t}(b_j^{t-1}, a_j^{t-1}, o_j^t, b_j^t)$ is replaced with the Kronecker delta function $\delta_K(APPEND(h_j^{t-1}, o_j^t), h_j^t)$.*

Above, $b_j^{t-1}$ and $b_j^t$ are the belief elements of $\theta_j^{t-1}$ and $\theta_j^t$, respectively, $\beta$ is a normalizing constant, and $Pr(a_j^{t-1}|\theta_j^{t-1})$ is the probability that $a_j^{t-1}$ is Bayesian rational for agent described by type $\theta_j^{t-1}$. This probability is equal to $\frac{1}{|OPT(\theta_j)|}$ if $a_j^{t-1} \in OPT(\theta_j)$, and it is equal to zero otherwise. We define $OPT$ in Section 5.2.[12] For the case of $j$'s subintentional model, $is = (s, m_j)$, $h_j^{t-1}$ and $h_j^t$ are the observation histories which are part of $m_j^{t-1}$, and $m_j^t$ respectively, $O_j$ is the observation function in $m_j^t$, and $Pr(a_j^{t-1}|m_j^{t-1})$ is the probability assigned by $m_j^{t-1}$ to $a_j^{t-1}$. APPEND returns a string with the second argument appended to the first. The proofs of the propositions are in the Appendix.

Proposition 2 and Eq. 6 have a lot in common with belief update in POMDPs, as should be expected. Both depend on agent $i$'s observation and transition functions. However, since agent $i$'s observations also depend on agent $j$'s actions, the probabilities of various actions of $j$ have to be included (in the first line of Eq. 6.) Further, since the update of agent $j$'s model depends on what $j$ observes, the probabilities of various observations of $j$ have to be included (in the second line of Eq. 6.) The update of $j$'s beliefs is represented by the $\tau_{\theta_j}$ term. The belief update can easily be generalized to the setting where more than one other agents co-exist with agent $i$.

12. If the agent's prior belief over $IS_i$ is given by a probability density function then the $\sum_{is^{t-1}}$ is replaced by an integral. In that case $\tau_{\theta_j^t}(b_j^{t-1}, a_j^{t-1}, o_j^t, b_j^t)$ takes the form of Dirac delta function over argument $b_j^{t-1}$: $\delta_D(SE_{\theta_j^t}(b_j^{t-1}, a_j^{t-1}, o_j^t) - b_j^t)$.

### 5.2 Value Function and Solutions in I-POMDPs

Analogously to POMDPs, each belief state in I-POMDP has an associated value reflecting the maximum payoff the agent can expect in this belief state:

$$U(\theta_i) = \max_{a_i \in A_i} \left\{ \sum_{is} ER_i(is, a_i) b_i(is) + \gamma \sum_{o_i \in \Omega_i} Pr(o_i|a_i, b_i) U(\langle SE_{\theta_i}(b_i, a_i, o_i), \widehat{\theta}_i \rangle) \right\} \tag{7}$$

where, $ER_i(is, a_i) = \sum_{a_j} R_i(is, a_i, a_j) Pr(a_j|m_j)$. Eq. 7 is a basis for value iteration in I-POMDPs.

Agent $i$'s optimal action, $a_i^*$, for the case of infinite horizon criterion with discounting, is an element of the set of optimal actions for the belief state, $OPT(\theta_i)$, defined as:

$$OPT(\theta_i) = \underset{a_i \in A_i}{argmax} \left\{ \sum_{is} ER_i(is, a_i) b_i(is) + \gamma \sum_{o_i \in \Omega_i} Pr(o_i|a_i, b_i) U(\langle SE_{\theta_i}(b_i, a_i, o_i), \widehat{\theta}_i \rangle) \right\} \tag{8}$$

As in the case of belief update, due to possibly infinitely nested beliefs, a step of value iteration and optimal actions are only asymptotically computable.

## 6. Finitely Nested I-POMDPs

Possible infinite nesting of agents' beliefs in intentional models presents an obvious obstacle to computing the belief updates and optimal solutions. Since the models of agents with infinitely nested beliefs correspond to agent functions which are not computable it is natural to consider finite nestings. We follow approaches in game theory (Aumann, 1999; Brandenburger & Dekel, 1993; Fagin et al., 1999), extend our previous work (Gmytrasiewicz & Durfee, 2000), and construct finitely nested I-POMDPs bottom-up. Assume a set of physical states of the world $S$, and two agents $i$ and $j$. Agent $i$'s 0-th level beliefs, $b_{i,0}$, are probability distributions over $S$. Its 0-th level types, $\Theta_{i,0}$, contain its 0-th level beliefs, and its frames, and analogously for agent $j$. 0-level types are, therefore, POMDPs.[13] 0-level models include 0-level types (i.e., intentional models) and the subintentional models, elements of $SM$. An agent's first level beliefs are probability distributions over physical states and 0-level models of the other agent. An agent's first level types consist of its first level beliefs and frames. Its first level models consist of the types upto level 1 and the subintentional models. Second level beliefs are defined in terms of first level models and so on. Formally, define spaces:

$$\begin{array}{lllllllll}
IS_{i,0} & = & S, & \Theta_{j,0} & = & \{\langle b_{j,0}, \widehat{\theta}_j \rangle : b_{j,0} \in \Delta(IS_{j,0})\}, & M_{j,0} & = & \Theta_{j,0} \cup SM_j \\
IS_{i,1} & = & S \times M_{j,0}, & \Theta_{j,1} & = & \{\langle b_{j,1}, \widehat{\theta}_j \rangle : b_{j,1} \in \Delta(IS_{j,1})\}, & M_{j,1} & = & \Theta_{j,1} \cup M_{j,0} \\
 & \vdots & & & \vdots & & & & \\
IS_{i,l} & = & S \times M_{j,l-1}, & \Theta_{j,l} & = & \{\langle b_{j,l}, \widehat{\theta}_j \rangle : b_{j,l} \in \Delta(IS_{j,l})\}, & M_{j,l} & = & \Theta_{j,l} \cup M_{j,l-1}
\end{array}$$

**Definition 4. (Finitely Nested I-POMDP)** *A finitely nested I-POMDP of agent $i$, I-POMDP$_{i,l}$, is:*

$$\textit{I-POMDP}_{i,l} = \langle IS_{i,l}, A, T_i, \Omega_i, O_i, R_i \rangle \tag{9}$$

13. In 0-level types the other agent's actions are folded into the $T$, $O$ and $R$ functions as noise.

The parameter $l$ will be called the *strategy level* of the finitely nested I-POMDP. The belief update, value function, and the optimal actions for finitely nested I-POMDPs are computed using Equation 6 and Equation 8, but recursion is guaranteed to terminate at 0-th level and subintentional models.

Agents which are more strategic are capable of modeling others at deeper levels (i.e., all levels up to their own strategy level $l$), but are always only boundedly optimal. As such, these agents could fail to predict the strategy of a more sophisticated opponent. The fact that the computability of an agent function implies that the agent may be suboptimal during interactions has been pointed out by Binmore (1990), and proved more recently by Nachbar and Zame (1996). Intuitively, the difficulty is that an agent's unbounded optimality would have to include the capability to model the other agent's modeling the original agent. This leads to an impossibility result due to self-reference, which is very similar to Gödel's incompleteness theorem and the halting problem (Brandenburger, 2002). On a positive note, some convergence results (Kalai & Lehrer, 1993) strongly suggest that approximate optimality is achievable, although their applicability to our work remains open.

As we mentioned, the 0-th level types are POMDPs. They provide probability distributions over actions of the agent modeled at that level to models with strategy level of 1. Given probability distributions over other agent's actions the level-1 models can themselves be solved as POMDPs, and provide probability distributions to yet higher level models. Assume that the number of models considered at each level is bound by a number, $M$. Solving an *I-POMDP*$_{i,l}$ in then equivalent to solving $O(M^l)$ POMDPs. Hence, the complexity of solving an *I-POMDP*$_{i,l}$ is PSPACE-hard for finite time horizons,[14] and undecidable for infinite horizons, just like for POMDPs.

### 6.1 Some Properties of I-POMDPs

In this section we establish two important properties, namely convergence of value iteration and piece-wise linearity and convexity of the value function, for finitely nested I-POMDPs.

#### 6.1.1 CONVERGENCE OF VALUE ITERATION

For an agent $i$ and its *I-POMDP*$_{i,l}$, we can show that the sequence of value functions, $\{U^n\}$, where $n$ is the horizon, obtained by value iteration defined in Eq. 7, converges to a unique fixed-point, $U^*$.

Let us define a *backup* operator $H : B \rightarrow B$ such that $U^n = HU^{n-1}$, and $B$ is the set of all bounded value functions. In order to prove the convergence result, we first establish some of the properties of $H$.

**Lemma 1 (Isotonicity).** *For any finitely nested I-POMDP value functions $V$ and $U$, if $V \leq U$, then $HV \leq HU$.*

The proof of this lemma is analogous to one due to Hauskrecht (1997), for POMDPs. It is also sketched in the Appendix. Another important property exhibited by the backup operator is the property of contraction.

**Lemma 2 (Contraction).** *For any finitely nested I-POMDP value functions $V$, $U$ and a discount factor $\gamma \in (0, 1)$, $||HV - HU|| \leq \gamma||V - U||$.*

The proof of this lemma is again similar to the corresponding one in POMDPs (Hausktecht, 1997). The proof makes use of Lemma 1. $||\cdot||$ is the supremum norm.

14. Usually PSPACE-complete since the number of states in I-POMDPs is likely to be larger than the time horizon (Papadimitriou & Tsitsiklis, 1987).

Under the contraction property of $H$, and noting that the space of value functions along with the supremum norm forms a complete normed space (Banach space), we can apply the Contraction Mapping Theorem (Stokey & Lucas, 1989) to show that value iteration for I-POMDPs converges to a unique fixed point (optimal solution). The following theorem captures this result.

**Theorem 1 (Convergence).** *For any finitely nested I-POMDP, the value iteration algorithm starting from any arbitrary well-defined value function converges to a unique fixed-point.*

The detailed proof of this theorem is included in the Appendix.

As in the case of POMDPs (Russell & Norvig, 2003), the error in the iterative estimates, $U^n$, for finitely nested I-POMDPs, i.e., $||U^n - U^*||$, is reduced by the factor of at least $\gamma$ on each iteration. Hence, the number of iterations, $N$, needed to reach an error of at most $\epsilon$ is:

$$N = \lceil \log(R_{max}/\epsilon(1-\gamma))/\log(1/\gamma) \rceil \tag{10}$$

where $R_{max}$ is the upper bound of the reward function.

#### 6.1.2 PIECEWISE LINEARITY AND CONVEXITY

Another property that carries over from POMDPs to finitely nested I-POMDPs is the piecewise linearity and convexity (PWLC) of the value function. Establishing this property allows us to decompose the I-POMDP value function into a set of *alpha* vectors, each of which represents a policy tree. The PWLC property enables us to work with sets of alpha vectors rather than perform value iteration over the continuum of agent's beliefs. Theorem 2 below states the PWLC property of the I-POMDP value function.

**Theorem 2 (PWLC).** *For any finitely nested I-POMDP, $U$ is piecewise linear and convex.*

The complete proof of Theorem 2 is included in the Appendix. The proof is similar to one due to Smallwood and Sondik (1973) for POMDPs and proceeds by induction. The basis case is established by considering the horizon 1 value function. Showing the PWLC for the inductive step requires substituting the belief update (Eq. 6) into Eq. 7, followed by factoring out the belief from both terms of the equation.

## 7. Example: Multi-agent Tiger Game

To illustrate optimal sequential behavior of agents in multi-agent settings we apply our I-POMDP framework to the multi-agent tiger game, a traditional version of which we described before.

### 7.1 Definition

Let us denote the actions of opening doors and listening as OR, OL and L, as before. TL and TR denote states corresponding to tiger located behind the left and right door, respectively. The transition, reward and observation functions depend now on the actions of both agents. Again, we assume that the tiger location is chosen randomly in the next time step if any of the agents opened any doors in the current step. We also assume that the agent hears the tiger's growls, GR and GL, with the accuracy of 85%. To make the interaction more interesting we added an observation of door creaks, which depend on the action executed by the other agent. Creak right, CR, is likely due to the other agent having opened the right door, and similarly for creak left, CL. Silence, S, is a good indication that the other agent did not open doors and listened instead. We assume that the accuracy of creaks is 90%. We also assume that the agent's payoffs are analogous to the single agent versions described in Section 3.2 to make these cases comparable. Note that the result of this assumption is that the other agent's actions do not impact the original agent's payoffs directly, but rather indirectly by resulting in states that matter to the original agent. Table 1 quantifies these factors.

| $\langle a_i, a_j \rangle$ | **State** | **TL** | **TR** |
|---|---|---|---|
| $\langle OL, * \rangle$ | * | 0.5 | 0.5 |
| $\langle OR, * \rangle$ | * | 0.5 | 0.5 |
| $\langle *, OL \rangle$ | * | 0.5 | 0.5 |
| $\langle *, OR \rangle$ | * | 0.5 | 0.5 |
| $\langle L, L \rangle$ | $TL$ | 1.0 | 0 |
| $\langle L, L \rangle$ | $TR$ | 0 | 1.0 |

Transition function: $T_i = T_j$

| $\langle a_i, a_j \rangle$ | **TL** | **TR** |
|---|---|---|
| $\langle OR, OR \rangle$ | 10 | -100 |
| $\langle OL, OL \rangle$ | -100 | 10 |
| $\langle OR, OL \rangle$ | 10 | -100 |
| $\langle OL, OR \rangle$ | -100 | 10 |
| $\langle L, L \rangle$ | -1 | -1 |
| $\langle L, OR \rangle$ | -1 | -1 |
| $\langle OR, L \rangle$ | 10 | -100 |
| $\langle L, OL \rangle$ | -1 | -1 |
| $\langle OL, L \rangle$ | -100 | 10 |

| $\langle a_i, a_j \rangle$ | **TL** | **TR** |
|---|---|---|
| $\langle OR, OR \rangle$ | 10 | -100 |
| $\langle OL, OL \rangle$ | -100 | 10 |
| $\langle OR, OL \rangle$ | -100 | 10 |
| $\langle OL, OR \rangle$ | 10 | -100 |
| $\langle L, L \rangle$ | -1 | -1 |
| $\langle L, OR \rangle$ | 10 | -100 |
| $\langle OR, L \rangle$ | -1 | -1 |
| $\langle L, OL \rangle$ | -100 | 10 |
| $\langle OL, L \rangle$ | -1 | -1 |

Reward functions of agents $i$ and $j$

| $\langle a_i, a_j \rangle$ | **State** | $\langle$ **GL, CL** $\rangle$ | $\langle$ **GL, CR** $\rangle$ | $\langle$ **GL, S** $\rangle$ | $\langle$ **GR, CL** $\rangle$ | $\langle$ **GR, CR** $\rangle$ | $\langle$ **GR, S** $\rangle$ |
|---|---|---|---|---|---|---|---|
| $\langle L, L \rangle$ | $TL$ | 0.85*0.05 | 0.85*0.05 | 0.85*0.9 | 0.15*0.05 | 0.15*0.05 | 0.15*0.9 |
| $\langle L, L \rangle$ | $TR$ | 0.15*0.05 | 0.15*0.05 | 0.15*0.9 | 0.85*0.05 | 0.85*0.05 | 0.85*0.9 |
| $\langle L, OL \rangle$ | $TL$ | 0.85*0.9 | 0.85*0.05 | 0.85*0.05 | 0.15*0.9 | 0.15*0.05 | 0.15*0.05 |
| $\langle L, OL \rangle$ | $TR$ | 0.15*0.9 | 0.15*0.05 | 0.15*0.05 | 0.85*0.9 | 0.85*0.05 | 0.85*0.05 |
| $\langle L, OR \rangle$ | $TL$ | 0.85*0.05 | 0.85*0.9 | 0.85*0.05 | 0.15*0.05 | 0.15*0.9 | 0.15*0.05 |
| $\langle L, OR \rangle$ | $TR$ | 0.15*0.05 | 0.15*0.9 | 0.15*0.05 | 0.85*0.05 | 0.85*0.9 | 0.85*0.05 |
| $\langle OL, * \rangle$ | $*$ | 1/6 | 1/6 | 1/6 | 1/6 | 1/6 | 1/6 |
| $\langle OR, * \rangle$ | $*$ | 1/6 | 1/6 | 1/6 | 1/6 | 1/6 | 1/6 |

| $\langle a_i, a_j \rangle$ | **State** | $\langle$ **GL, CL** $\rangle$ | $\langle$ **GL, CR** $\rangle$ | $\langle$ **GL, S** $\rangle$ | $\langle$ **GR, CL** $\rangle$ | $\langle$ **GR, CR** $\rangle$ | $\langle$ **GR, S** $\rangle$ |
|---|---|---|---|---|---|---|---|
| $\langle L, L \rangle$ | $TL$ | 0.85*0.05 | 0.85*0.05 | 0.85*0.9 | 0.15*0.05 | 0.15*0.05 | 0.15*0.9 |
| $\langle L, L \rangle$ | $TR$ | 0.15*0.05 | 0.15*0.05 | 0.15*0.9 | 0.85*0.05 | 0.85*0.05 | 0.85*0.9 |
| $\langle OL, L \rangle$ | $TL$ | 0.85*0.9 | 0.85*0.05 | 0.85*0.05 | 0.15*0.9 | 0.15*0.05 | 0.15*0.05 |
| $\langle OL, L \rangle$ | $TR$ | 0.15*0.9 | 0.15*0.05 | 0.15*0.05 | 0.85*0.9 | 0.85*0.05 | 0.85*0.05 |
| $\langle OR, L \rangle$ | $TL$ | 0.85*0.05 | 0.85*0.9 | 0.85*0.05 | 0.15*0.05 | 0.15*0.9 | 0.15*0.05 |
| $\langle OR, L \rangle$ | $TR$ | 0.15*0.05 | 0.15*0.9 | 0.15*0.05 | 0.85*0.05 | 0.85*0.9 | 0.85*0.05 |
| $\langle *, OL \rangle$ | $*$ | 1/6 | 1/6 | 1/6 | 1/6 | 1/6 | 1/6 |
| $\langle *, OR \rangle$ | $*$ | 1/6 | 1/6 | 1/6 | 1/6 | 1/6 | 1/6 |

Observation functions of agents $i$ and $j$.

Table 1: Transition, reward, and observation functions for the multi-agent Tiger game.

When an agent makes its choice in the multi-agent tiger game, it considers what it believes about the location of the tiger, as well as whether the other agent will listen or open a door, which in turn depends on the other agent's beliefs, reward function, optimality criterion, etc.[15] In particular, if the other agent were to open any of the doors the tiger location in the next time step would be chosen randomly. Thus, the information obtained from hearing the previous growls would have to be discarded. We simplify the situation by considering $i$'s I-POMDP with a single level of nesting, assuming that all of the agent $j$'s properties, except for beliefs, are known to $i$, and that $j$'s time horizon is equal to $i$'s. In other words, $i$'s uncertainty pertains only to $j$'s beliefs and not to its frame. Agent $i$'s interactive state space is, $IS_{i,1} = S \times \Theta_{j,0}$, where $S$ is the physical state, $S$={TL,

15. We assume an intentional model of the other agent here.

TR}, and $\Theta_{j,0}$ is a set of intentional models of agent $j$'s, each of which differs only in $j$'s beliefs over the location of the tiger.

### 7.2 Examples of the Belief Update

In Section 5, we presented the belief update equation for I-POMDPs (Eq. 6). Here we consider examples of beliefs, $b_{i,1}$, of agent $i$, which are probability distributions over $S \times \Theta_{j,0}$. Each 0-th level type of agent $j$, $\theta_{j,0} \in \Theta_{j,0}$, contains a "flat" belief as to the location of the tiger, which can be represented by a single probability assignment – $b_{j,0} = p_j(TL)$.

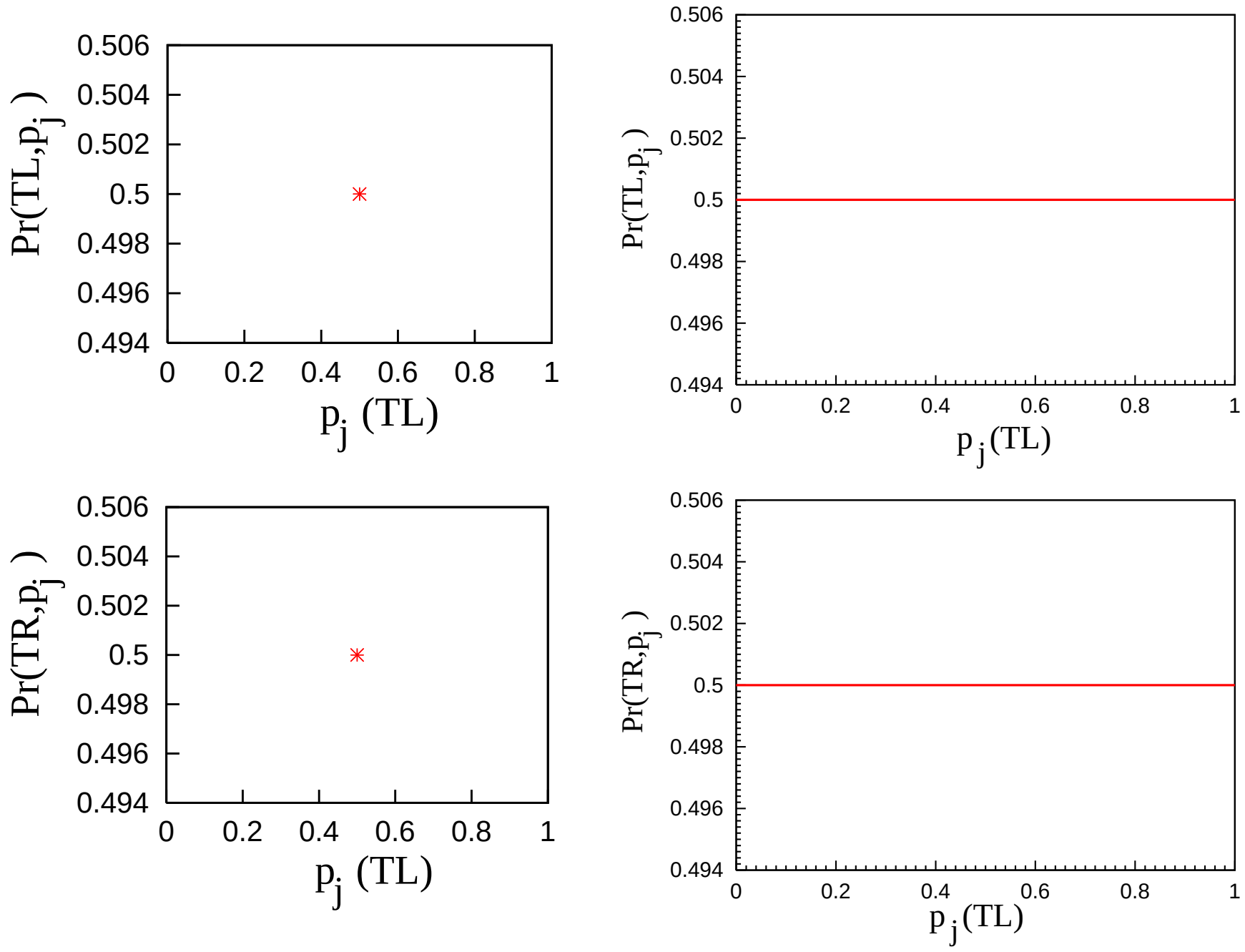

$(i)$ $(ii)$

Figure 5: Two examples of singly nested belief states of agent $i$. In each case $i$ has no information about the tiger's location. In $(i)$ agent $i$ knows that $j$ does not know the location of the tiger; the single point (star) denotes a Dirac delta function which integrates to the height of the point, here 0.5 . In $(ii)$ agent $i$ is uninformed about $j$'s beliefs about tiger's location.

In Fig. 5 we show some examples of level 1 beliefs of agent $i$. In each case $i$ does not know the location of the tiger so that the marginals in the top and bottom sections of the figure sum up to 0.5 for probabilities of TL and TR each. In Fig. 5$(i)$, $i$ knows that $j$ assigns 0.5 probability to tiger being behind the left door. This is represented using a Dirac delta function. In Fig. 5$(ii)$, agent $i$ is uninformed about $j$'s beliefs. This is represented as a uniform probability density over all values of the probability $j$ could assign to state TL.

To make the presentation of the belief update more transparent we decompose the formula in Eq. 6 into two steps:

- *Prediction:* When agent $i$ performs an action $a_i^{t-1}$, and given that agent $j$ performs $a_j^{t-1}$, the predicted belief state is:

$$\widehat{b}_i^t(is^t) = Pr(is^t|a_i^{t-1}, a_j^{t-1}, b_i^{t-1}) \quad = \sum_{is^{t-1}|\widehat{\theta}_j^{t-1}=\widehat{\theta}_j^t} b_i^{t-1}(is^{t-1}) Pr(a_j^{t-1}|\theta_j^{t-1}) \times T(s^{t-1}, a^{t-1}, s^t) \sum_{o_j^t} O_j(s^t, a^{t-1}, o_j^t) \times \tau_{\theta_j^t}(b_j^{t-1}, a_j^{t-1}, o_j^t, b_j^t) \tag{11}$$

- *Correction:* When agent $i$ perceives an observation, $o_i^t$, the predicted belief states, $Pr(\cdot|a_i^{t-1}, a_j^{t-1}, b_i^{t-1})$, are combined according to:

$$b_i^t(is^t) = Pr(is^t|o_i^t, a_i^{t-1}, b_i^{t-1}) = \beta \sum_{a_j^{t-1}} O_i(s^t, a^{t-1}, o_i^t) Pr(is^t|a_i^{t-1}, a_j^{t-1}, b_i^{t-1}) \tag{12}$$

where $\beta$ is the normalizing constant.

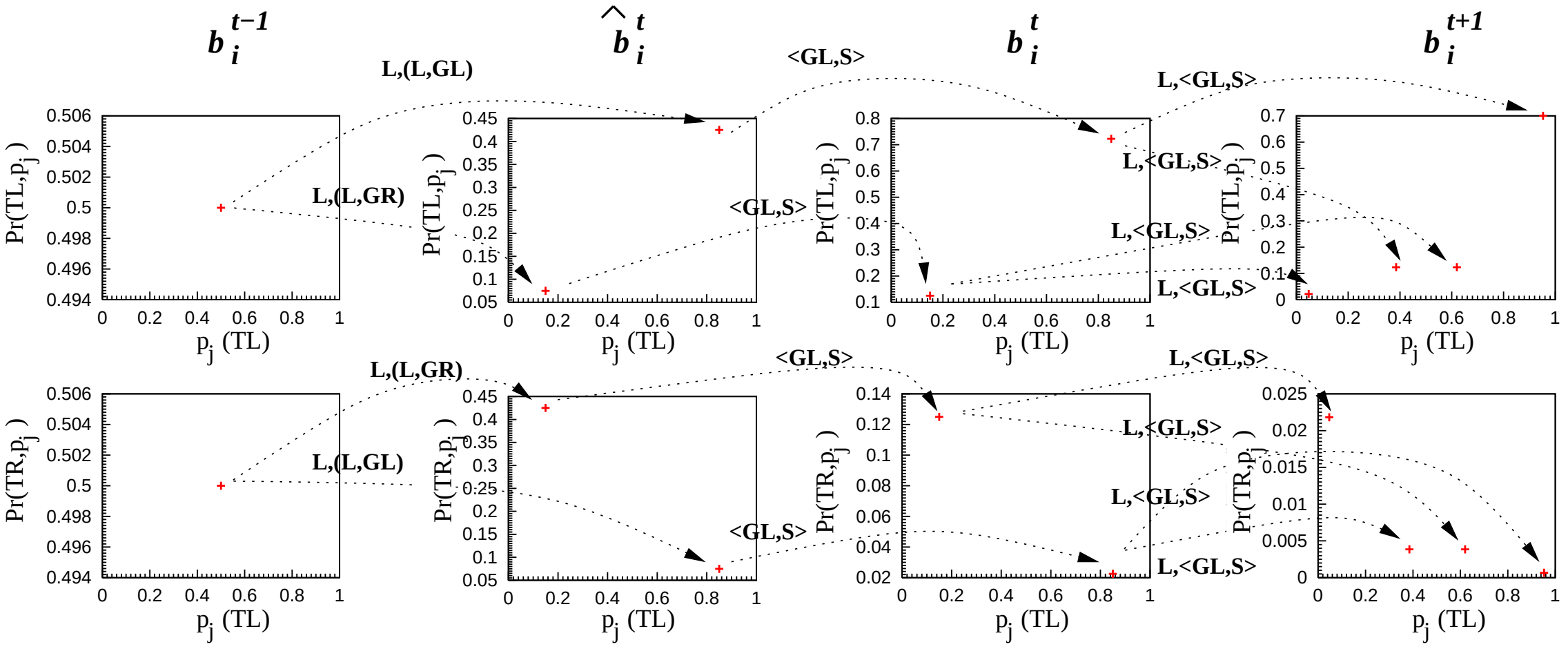

Figure 6: A trace of the belief update of agent $i$. $(a)$ depicts the prior. $(b)$ is the result of prediction given $i$'s listening action, L, and a pair denoting $j$'s action and observation. $i$ knows that $j$ will listen and could hear tiger's growl on the right or the left, and that the probabilities $j$ would assign to TL are 0.15 or 0.85, respectively. $(c)$ is the result of correction after $i$ observes tiger's growl on the left and no creaks, ⟨GL,S⟩. The probability $i$ assigns to TL is now greater than TR. $(d)$ depicts the results of another update (both prediction and correction) after another listen action of $i$ and the same observation, ⟨GL,S⟩.

Each discrete point above denotes, again, a Dirac delta function which integrates to the height of the point.

In Fig. 6, we display the example trace through the update of singly nested belief. In the first column of Fig. 6, labeled (a), is an example of agent $i$'s prior belief we introduced before, according to which $i$ knows that $j$ is uninformed of the location of the tiger.[16] Let us assume that $i$ listens and hears a growl from the left and no creaks. The second column of Fig. 6, (b), displays the *predicted* belief after $i$ performs the listen action (Eq. 11). As part of the prediction step, agent $i$ must solve $j$'s model to obtain $j$'s optimal action when its belief is 0.5 (term $Pr(a_j^{t-1}|\theta_j)$ in Eq. 11). Given the value function in Fig. 3, this evaluates to probability of 1 for listen action, and zero for opening of any of the doors. $i$ also updates $j$'s belief given that $j$ listens and hears the tiger growling from either the left, GL, or right, GR, (term $\tau_{\theta_j^t}(b_j^{t-1}, a_j^{t-1}, o_j^t, b_j^t)$ in Eq. 11). Agent $j$'s updated probabilities for tiger being on the left are 0.85 and 0.15, for $j$'s hearing GL and GR, respectively. If the tiger is on the left (top of Fig. 6 (b)) $j$'s observation GL is more likely, and consequently $j$'s assigning the probability of 0.85 to state TL is more likely ($i$ assigns a probability of 0.425 to this state.) When the tiger is on the right $j$ is more likely to hear GR and $i$ assigns the lower probability, 0.075, to $j$'s assigning a probability 0.85 to tiger being on the left. The third column, (c), of Fig. 6 shows the posterior belief after the *correction* step. The belief in column (b) is updated to account for $i$'s hearing a growl from the left and no creaks, $\langle$GL,S$\rangle$. The resulting marginalised probability of the tiger being on the left is higher (0.85) than that of the tiger being on the right. If we assume that in the next time step $i$ again listens and hears the tiger growling from the left and no creaks, the belief state depicted in the fourth column of Fig. 6 results.

In Fig. 7 we show the belief update starting from the prior in Fig. 5 $(ii)$, according to which agent $i$ initially has no information about what $j$ believes about the tiger's location.

The traces of belief updates in Fig. 6 and Fig. 7 illustrate the changing state of information agent $i$ has about the other agent's beliefs. The benefit of representing these updates explicitly is that, at each stage, $i$'s optimal behavior depends on its estimate of probabilities of $j$'s actions. The more informative these estimates are the more value agent $i$ can expect out of the interaction. Below, we show the increase in the value function for I-POMDPs compared to POMDPs with the noise factor.

### 7.3 Examples of Value Functions

This section compares value functions obtained from solving a POMDP with a static noise factor, accounting for the presence of another agent,[17] to value functions of level-1 I-POMDP. The advantage of more refined modeling and update in I-POMDPs is due to two factors. First is the ability to keep track of the other agent's state of beliefs to better predict its future actions. The second is the ability to adjust the other agent's time horizon as the number of steps to go during the interaction decreases. Neither of these is possible within the classical POMDP formalism.

We continue with the simple example of *I-POMDP*$_{i,1}$ of agent $i$. In Fig. 8 we display $i$'s value function for the time horizon of 1, assuming that $i$'s initial belief as to the value $j$ assigns to TL, $p_j(TL)$, is as depicted in Fig. 5 $(ii)$, i.e. $i$ has no information about what $j$ believes about tiger's location. This value function is identical to the value function obtained for an agent using a traditional POMDP framework with noise, as well as single agent POMDP which we described in Section 3.2. The value functions overlap since agents do not have to update their beliefs and the advantage of more refined modeling of agent $j$ in $i$'s I-POMDP does not become apparent. Put another way, when agent $i$ models $j$ using an intentional model, it concludes that agent $j$ will open each door with probability 0.1 and listen with probability 0.8. This coincides with the noise factor we described in Section 3.2.

16. The points in Fig. 7 again denote Dirac delta functions which integrate to the value equal to the points' height.
17. The POMDP with noise is the same as level-0 I-POMDP.

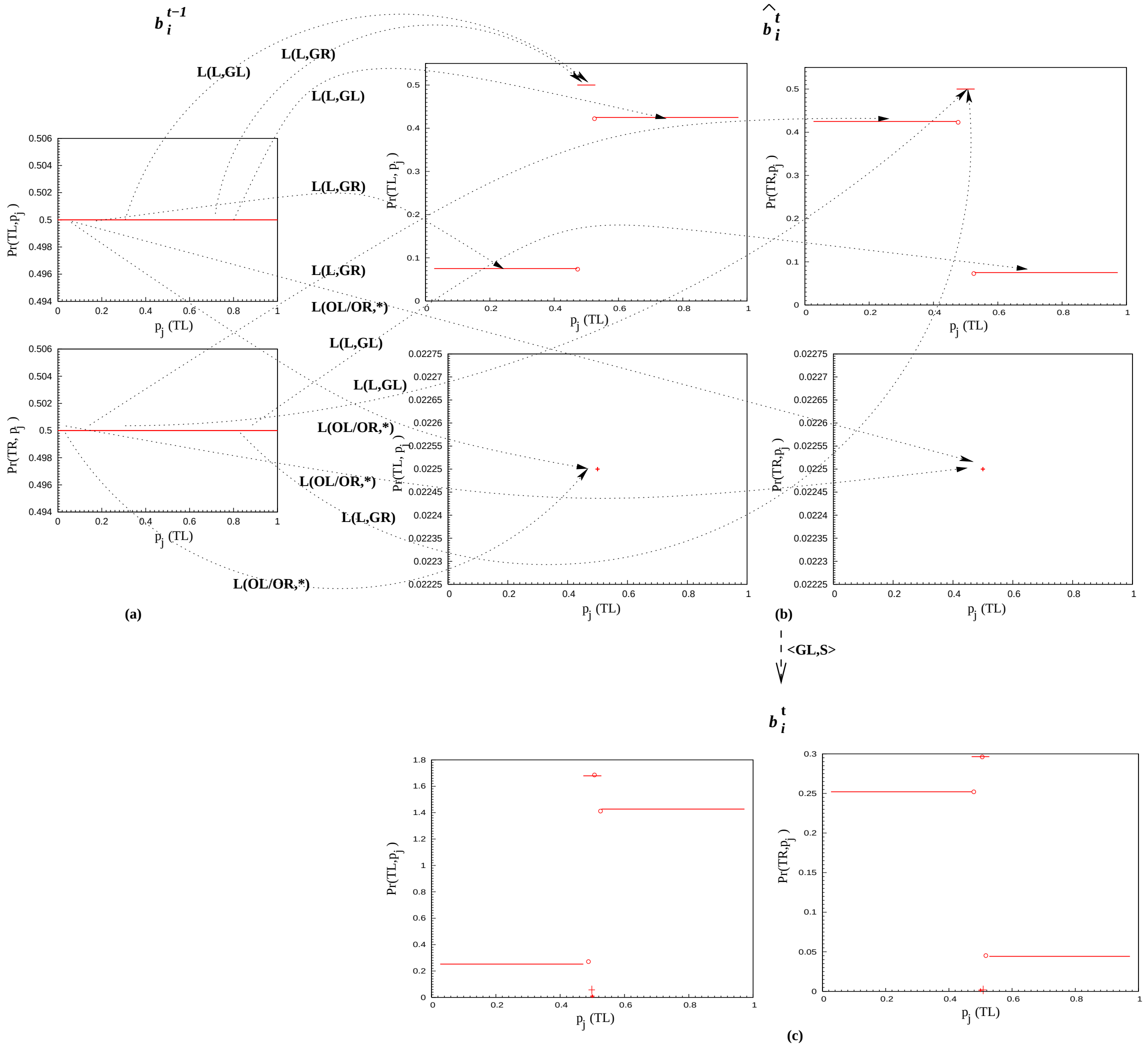

Figure 7: A trace of the belief update of agent $i$. $(a)$ depicts the prior according to which $i$ is uninformed about $j$'s beliefs. $(b)$ is the result of the prediction step after $i$'s listening action (L). The top half of $(b)$ shows $i$'s belief after it has listened and given that $j$ also listened. The two observations $j$ can make, GL and GR, each with probability dependent on the tiger's location, give rise to flat portions representing what $i$ knows about $j$'s belief in each case. The increased probability $i$ assigns to $j$'s belief between 0.472 and 0.528 is due to $j$'s updates after it hears GL and after it hears GR resulting in the same values in this interval. The bottom half of $(b)$ shows $i$'s belief after $i$ has listened and $j$ has opened the left or right door (plots are identical for each action and only one of them is shown). $i$ knows that $j$ has no information about the tiger's location in this case. $(c)$ is the result of correction after $i$ observes tiger's growl on the left and no creaks $\langle$GL,S$\rangle$. The plots in $(c)$ are obtained by performing a weighted summation of the plots in $(b)$. The probability $i$ assigns to TL is now greater than TR, and information about $j$'s beliefs allows $i$ to refine its prediction of $j$'s action in the next time step.

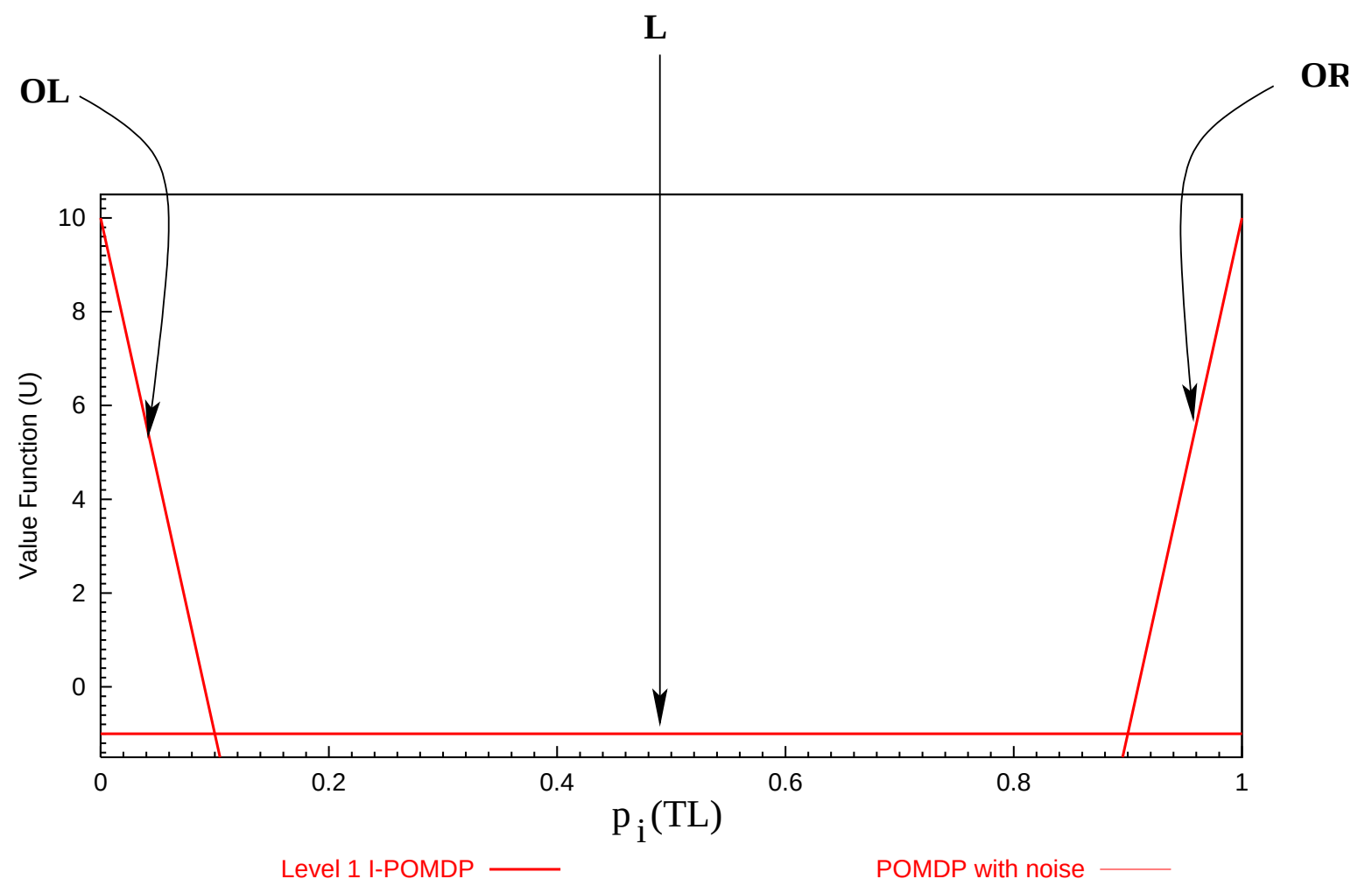

Figure 8: For time horizon of 1 the value functions obtained from solving a singly nested I-POMDP and a POMDP with noise factor overlap.

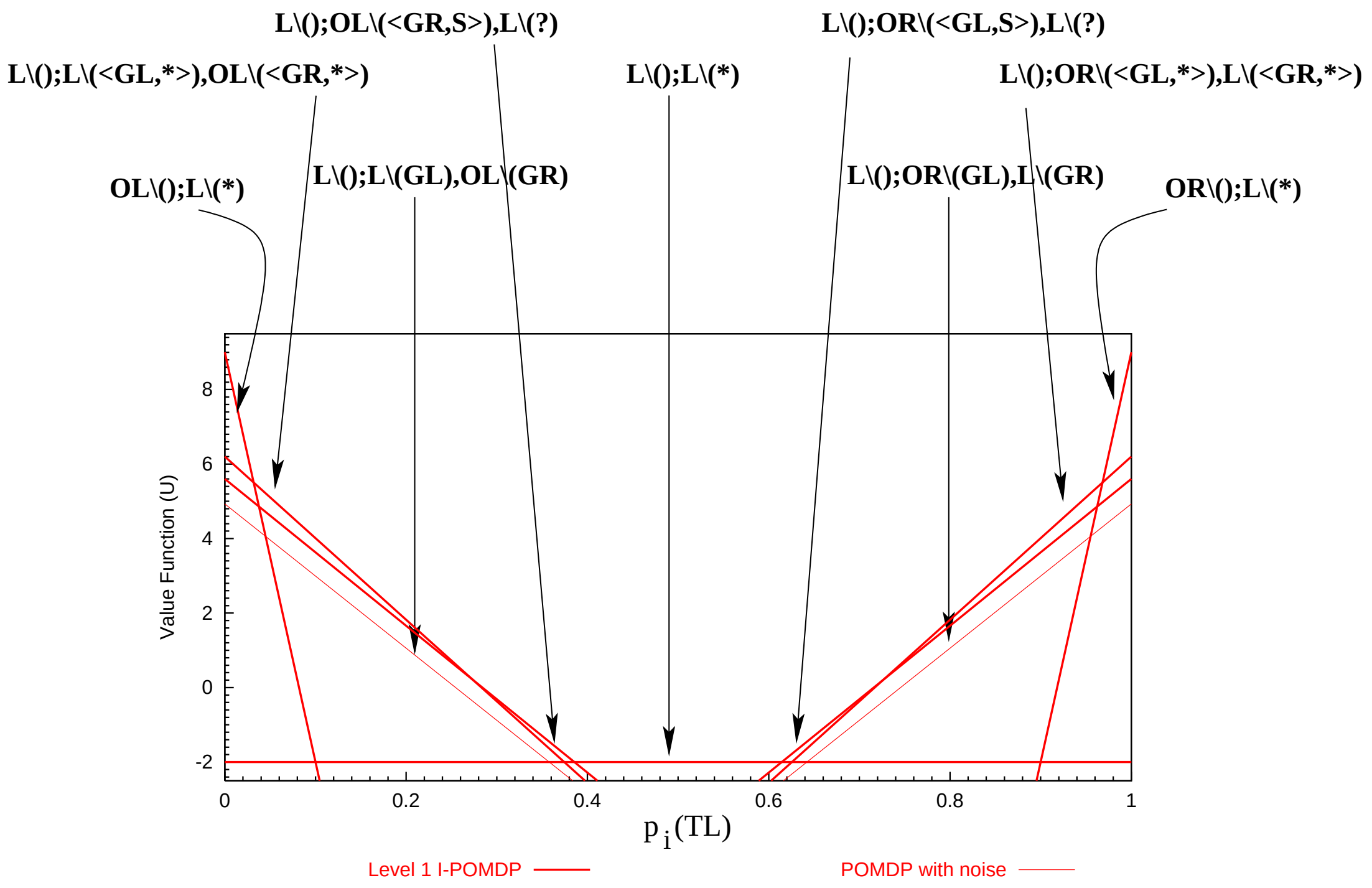

Figure 9: Comparison of value functions obtained from solving an I-POMDP and a POMDP with noise for time horizon of 2. I-POMDP value function dominates due to agent $i$ adjusting the behavior of agent $j$ to the remaining steps to go in the interaction.

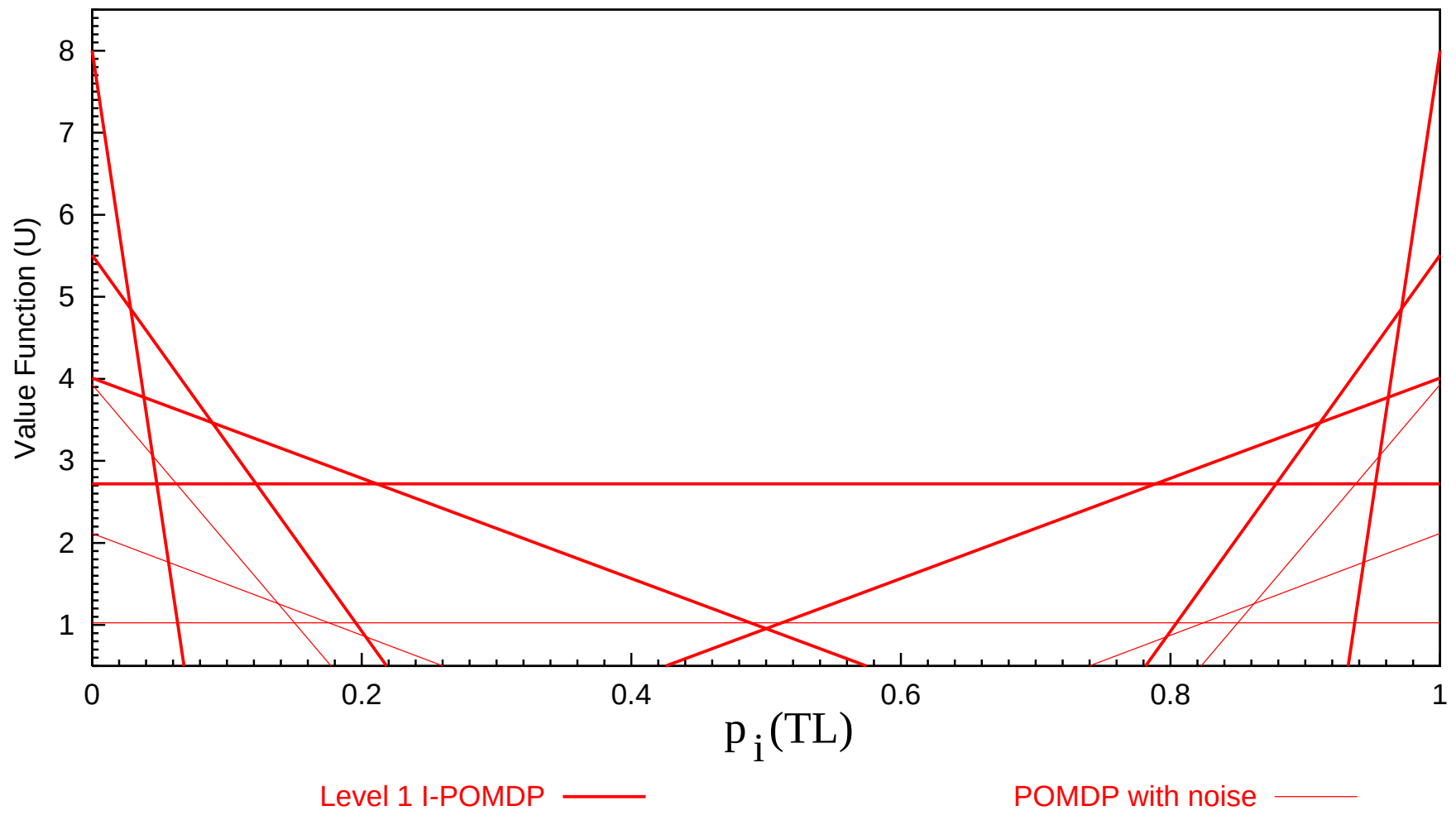

Figure 10: Comparison of value functions obtained from solving an I-POMDP and a POMDP with noise for time horizon of 3. I-POMDP value function dominates due to agent $i$'s adjusting $j$'s remaining steps to go, and due to $i$'s modeling $j$'s belief update. Both factors allow for better predictions of $j$'s actions during interaction. The descriptions of individual policies were omitted for clarity; they can be read off of Fig. 11.

In Fig. 9 we display $i$'s value functions for the time horizon of 2. The value function of *I-POMDP*$_{i,1}$ is higher than the value function of a POMDP with a noise factor. The reason is not related to the advantages of modeling agent $j$'s beliefs – this effect becomes apparent at the time horizon of 3 and longer. Rather, the I-POMDP solution dominates due to agent $i$ modeling $j$'s time horizon during interaction: $i$ knows that at the last time step $j$ will behave according to its optimal policy for time horizon of 1, while with two steps to go $j$ will optimize according to its 2 steps to go policy. As we mentioned, this effect cannot be modeled using a POMDP with a static noise factor included in the transition function.

Fig. 10 shows a comparison between the I-POMDP and the noisy POMDP value functions for horizon 3. The advantage of more refined agent modeling within the I-POMDP framework has increased.[18] Both factors, $i$'s adjusting $j$'s steps to go and $i$'s modeling $j$'s belief update during interaction are responsible for the superiority of values achieved using the I-POMDP. In particular, recall that at the second time step $i$'s information as to $j$'s beliefs about the tiger's location is as depicted in Fig. 7 (c). This enables $i$ to make a high quality prediction that, with two steps left to go, $j$ will perform its actions OL, L, and OR with probabilities 0.009076, 0.96591 and 0.02501, respectively (recall that for POMDP with noise these probabilities remained unchanged at 0.1, 0,8, and 0.1, respectively.)

Fig. 11 shows agent $i$'s policy graph for time horizon of 3. As usual, it prescribes the optimal first action depending on the initial belief as to the tiger's location. The subsequent actions depend on the observations received. The observations include creaks that are indicative of the other agent's

18. Note that I-POMDP solution is not as good as the solution of a POMDP for an agent operating alone in the environment shown in Fig. 3.

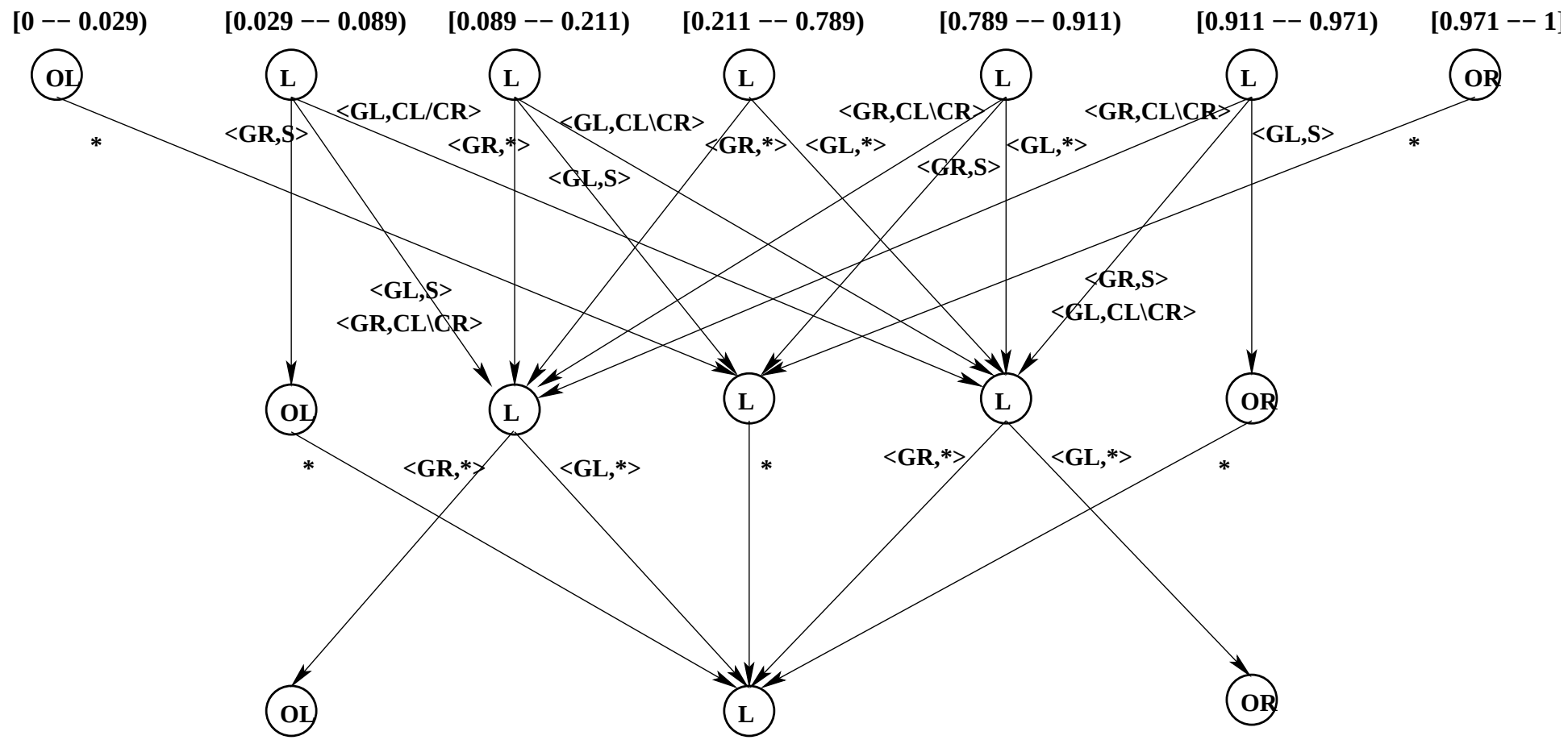

Figure 11: The policy graph corresponding to the I-POMDP value function in Fig. 10.

having opened a door. The creaks contain valuable information and allow the agent to make more refined choices, compared to ones in the noisy POMDP in Fig. 4. Consider the case when agent $i$ starts out with fairly strong belief as to the tiger's location, decides to listen (according to the four off-center top row "L" nodes in Fig. 11) and hears a door creak. The agent is then in the position to open either the left or the right door, even if that is counter to its initial belief. The reason is that the creak is an indication that the tiger's position has likely been reset by agent $j$ and that $j$ will then not open any of the doors during the following two time steps. Now, two growls coming from the same door lead to enough confidence to open the other door. This is because the agent $i$'s hearing of tiger's growls are indicative of the tiger's position in the state following the agents' actions,

Note that the value functions and the policy above depict a special case of agent $i$ having no information as to what probability $j$ assigns to tiger's location (Fig. 5 $(ii)$). Accounting for and visualizing all possible beliefs $i$ can have about $j$'s beliefs is difficult due to the complexity of the space of interactive beliefs. As our ongoing work indicates, a drastic reduction in complexity is possible without loss of information, and consequently representation of solutions in a manageable number of dimensions is indeed possible. We will report these results separately.

## 8. Conclusions

We proposed a framework for optimal sequential decision-making suitable for controlling autonomous agents interacting with other agents within an uncertain environment. We used the normative paradigm of decision-theoretic planning under uncertainty formalized as partially observable Markov decision processes (POMDPs) as a point of departure. We extended POMDPs to cases of agents interacting with other agents by allowing them to have beliefs not only about the physical environment, but also about the other agents. This could include beliefs about the others' abilities, sensing capabilities, beliefs, preferences, and intended actions. Our framework shares numerous properties with POMDPs, has analogously defined solutions, and reduces to POMDPs when agents are alone in the environment.

In contrast to some recent work on DEC-POMDPs (Bernstein et al., 2002; Nair et al., 2003), and to work motivated by game-theoretic equilibria (Boutilier, 1999; Hu & Wellman, 1998; Koller

& Milch, 2001; Littman, 1994), our approach is subjective and amenable to agents independently computing their optimal solutions.

The line of work presented here opens an area of future research on integrating frameworks for sequential planning with elements of game theory and Bayesian learning in interactive settings. In particular, one of the avenues of our future research centers on proving further formal properties of I-POMDPs, and establishing clearer relations between solutions to I-POMDPs and various flavors of equilibria. Another concentrates on developing efficient approximation techniques for solving I-POMDPs. As for POMDPs, development of approximate approaches to I-POMDPs is crucial for moving beyond toy problems. One promising approximation technique we are working on is particle filtering. We are also devising methods for representing I-POMDP solutions without assumptions about what's believed about other agents' beliefs. As we mentioned, in spite of the complexity of the interactive state space, there seem to be intuitive representations of belief partitions corresponding to optimal policies, analogous to those for POMDPs. Other research issues include the suitable choice of priors over models,[19] and the ways to fulfill the absolute continuity condition needed for convergence of probabilities assigned to the alternative models during interactions (Kalai & Lehrer, 1993).

## Acknowledgments

This research is supported by the National Science Foundation CAREER award IRI-9702132, and NSF award IRI-0119270.

## Appendix A. Proofs

*Proof of Propositions 1 and 2.* We start with Proposition 2, by applying the Bayes Theorem:

$$
\begin{aligned}
b_i^t(is^t) &= Pr(is^t|o_i^t, a_i^{t-1}, b_i^{t-1}) = \frac{Pr(is^t, o_i^t|a_i^{t-1}, b_i^{t-1})}{Pr(o_i^t|a_i^{t-1}, b_i^{t-1})} \\
&= \beta \sum_{is^{t-1}} b_i^{t-1}(is^{t-1}) Pr(is^t, o_i^t|a_i^{t-1}, is^{t-1}) \\
&= \beta \sum_{is^{t-1}} b_i^{t-1}(is^{t-1}) \sum_{a_j^{t-1}} Pr(is^t, o_i^t|a_i^{t-1}, a_j^{t-1}, is^{t-1}) Pr(a_j^{t-1}|a_i^{t-1}, is^{t-1}) \\
&= \beta \sum_{is^{t-1}} b_i^{t-1}(is^{t-1}) \sum_{a_j^{t-1}} Pr(is^t, o_i^t|a_i^{t-1}, a_j^{t-1}, is^{t-1}) Pr(a_j^{t-1}|is^{t-1}) \\
&= \beta \sum_{is^{t-1}} b_i^{t-1}(is^{t-1}) \sum_{a_j^{t-1}} Pr(a_j^{t-1}|m_j^{t-1}) Pr(o_t^i|is^t, a^{t-1}, is^{t-1}) Pr(is^t|a^{t-1}, is^{t-1}) \\
&= \beta \sum_{is^{t-1}} b_i^{t-1}(is^{t-1}) \sum_{a_j^{t-1}} Pr(a_j^{t-1}|m_j^{t-1}) Pr(o_t^i|is^t, a^{t-1}) Pr(is^t|a^{t-1}, is^{t-1}) \\
&= \beta \sum_{is^{t-1}} b_i^{t-1}(is^{t-1}) \sum_{a_j^{t-1}} Pr(a_j^{t-1}|m_j^{t-1}) O_i(s^t, a^{t-1}, o_i^t) Pr(is^t|a^{t-1}, is^{t-1})
\end{aligned}
\quad (13)
$$

19. We are looking at Kolmogorov complexity (Li & Vitanyi, 1997) as a possible way to assign priors.

To simplify the term $Pr(is^t|a^{t-1}, is^{t-1})$ let us substitute the interactive state $is^t$ with its components. When $m_j$ in the interactive states is intentional: $is^t = (s^t, \theta^t_j) = (s^t, b^t_j, \widehat{\theta}^t_j)$.

$$\begin{aligned} Pr(is^t|a^{t-1}, is^{t-1}) &= Pr(s^t, b^t_j, \widehat{\theta}^t_j|a^{t-1}, is^{t-1}) \\ &= Pr(b^t_j|s^t, \widehat{\theta}^t_j, a^{t-1}, is^{t-1})Pr(s^t, \widehat{\theta}^t_j|a^{t-1}, is^{t-1}) \\ &= Pr(b^t_j|s^t, \widehat{\theta}^t_j, a^{t-1}, is^{t-1})Pr(\widehat{\theta}^t_j|s^t, a^{t-1}, is^{t-1})Pr(s^t|a^{t-1}, is^{t-1}) \\ &= Pr(b^t_j|s^t, \widehat{\theta}^t_j, a^{t-1}, is^{t-1})I(\widehat{\theta}^{t-1}_j, \widehat{\theta}^t_j)T_i(s^{t-1}, a^{t-1}, s^t) \end{aligned} \tag{14}$$

When $m_j$ is subintentional: $is^t = (s^t, m^t_j) = (s^t, h^t_j, \widehat{m}^t_j)$.

$$\begin{aligned} Pr(is^t|a^{t-1}, is^{t-1}) &= Pr(s^t, h^t_j, \widehat{m}^t_j|a^{t-1}, is^{t-1}) \\ &= Pr(h^t_j|s^t, \widehat{m}^t_j, a^{t-1}, is^{t-1})Pr(s^t, \widehat{m}^t_j|a^{t-1}, is^{t-1}) \\ &= Pr(h^t_j|s^t, \widehat{m}^t_j, a^{t-1}, is^{t-1})Pr(\widehat{\theta}^t_j|s^t, a^{t-1}, is^{t-1})Pr(s^t|a^{t-1}, is^{t-1}) \\ &= Pr(h^t_j|s^t, \widehat{m}^t_j, a^{t-1}, is^{t-1})I(\widehat{m}^{t-1}_j, \widehat{m}^t_j)T_i(s^{t-1}, a^{t-1}, s^t) \end{aligned} \tag{14'}$$

The joint action pair, $a^{t-1}$, may change the physical state. The third term on the right-hand side of Eqs. 14 and $14'$ above captures this transition. We utilized the MNM assumption to replace the second terms of the equations with boolean identity functions, $I(\widehat{\theta}^{t-1}_j, \widehat{\theta}^t_j)$ and $I(\widehat{m}^{t-1}_j, \widehat{m}^t_j)$ respectively, which equal 1 if the two frames are identical, and 0 otherwise. Let us turn our attention to the first terms. If $m_j$ in $is^t$ and $is^{t-1}$ is intentional:

$$\begin{aligned} Pr(b^t_j|s^t, \widehat{\theta}^t_j, a^{t-1}, is^{t-1}) &= \textstyle\sum_{o^t_j} Pr(b^t_j|s^t, \widehat{\theta}^t_j, a^{t-1}, is^{t-1}, o^t_j)Pr(o^t_j|s^t, \widehat{\theta}^t_j, a^{t-1}, is^{t-1}) \\ &= \textstyle\sum_{o^t_j} Pr(b^t_j|s^t, \widehat{\theta}^t_j, a^{t-1}, is^{t-1}, o^t_j)Pr(o^t_j|s^t, \widehat{\theta}^t_j, a^{t-1}) \\ &= \textstyle\sum_{o^t_j} \tau_{\theta^t_j}(b^{t-1}_j, a^{t-1}_j, o^t_j, b^t_j)O_j(s_t, a^{t-1}, o^t_j) \end{aligned} \tag{15}$$

Else if it is subintentional:

$$\begin{aligned} Pr(h^t_j|s^t, \widehat{m}^t_j, a^{t-1}, is^{t-1}) &= \textstyle\sum_{o^t_j} Pr(h^t_j|s^t, \widehat{m}^t_j, a^{t-1}, is^{t-1}, o^t_j)Pr(o^t_j|s^t, \widehat{m}^t_j, a^{t-1}, is^{t-1}) \\ &= \textstyle\sum_{o^t_j} Pr(h^t_j|s^t, \widehat{m}^t_j, a^{t-1}, is^{t-1}, o^t_j)Pr(o^t_j|s^t, \widehat{m}^t_j, a^{t-1}) \\ &= \textstyle\sum_{o^t_j} \delta_K(\text{APPEND}(h^{t-1}_j, o^t_j), h^t_j)O_j(s_t, a^{t-1}, o^t_j) \end{aligned} \tag{15'}$$

In Eq. 15, the first term on the right-hand side is 1 if agent $j$'s belief update, $SE_{\theta_j}(b^{t-1}_j, a^{t-1}_j, o^t_j)$ generates a belief state equal to $b^t_j$. Similarly, in Eq. $15'$, the first term is 1 if appending the $o^t_j$ to $h^{t-1}_j$ results in $h^t_j$. $\delta_K$ is the Kronecker delta function. In the second terms on the right-hand side of the equations, the MNO assumption makes it possible to replace $Pr(o^t_j|s^t, \widehat{\theta}^t_j, a^{t-1})$ with $O_j(s^t, a^{t-1}, o^t_j)$, and $Pr(o^t_j|s^t, \widehat{m}^t_j, a^{t-1})$ with $O_j(s^t, a^{t-1}, o^t_j)$ respectively.

Let us now substitute Eq. 15 into Eq. 14.

$$Pr(is^t|a^{t-1}, is^{t-1}) = \textstyle\sum_{o^t_j} \tau_{\theta^t_j}(b^{t-1}_j, a^{t-1}_j, o^t_j, b^t_j)O_j(s^t, a^{t-1}, o^t_j)I(\widehat{\theta}^{t-1}_j, \widehat{\theta}^t_j)T_i(s^{t-1}, a^{t-1}, s^t) \tag{16}$$

Substituting Eq. $15'$ into Eq. $14'$ we get,

$$\begin{aligned} Pr(is^t|a^{t-1}, is^{t-1}) &= \textstyle\sum_{o^t_j} \delta_K(\text{APPEND}(h^{t-1}_j, o^t_j), h^t_j)O_j(s^t, a^{t-1}, o^t_j)I(\widehat{m}^{t-1}_j, \widehat{m}^t_j) \\ &\times T_i(s^{t-1}, a^{t-1}, s^t) \end{aligned} \tag{16'}$$

Replacing Eq. 16 into Eq. 13 we get:

$$\begin{aligned} b_i^t(is^t) &= \beta \sum_{is^{t-1}} b_i^{t-1}(is^{t-1}) \sum_{a_j^{t-1}} Pr(a_j^{t-1}|\theta_j^{t-1}) O_i(s^t, a^{t-1}, o_i^t) \sum_{o_j^t} \tau_{\theta_j^t}(b_j^{t-1}, a_j^{t-1}, o_j^t, b_j^t) \\ &\times O_j(s^t, a^{t-1}, o_j^t) I(\widehat{\theta}_j^{t-1}, \widehat{\theta}_j^t) T_i(s^{t-1}, a^{t-1}, s^t) \end{aligned} \tag{17}$$

Similarly, replacing Eq. 16′ into Eq. 13 we get:

$$\begin{aligned} b_i^t(is^t) &= \beta \sum_{is^{t-1}} b_i^{t-1}(is^{t-1}) \sum_{a_j^{t-1}} Pr(a_j^{t-1}|m_j^{t-1}) O_i(s^t, a^{t-1}, o_i^t) \\ &\times \sum_{o_j^t} \delta_K(\text{APPEND}(h_j^{t-1}, o_j^t), h_j^t) O_j(s^t, a^{t-1}, o_j^t) I(\widehat{m}_j^{t-1}, \widehat{m}_j^t) T_i(s^{t-1}, a^{t-1}, s^t) \end{aligned} \tag{17′}$$

We arrive at the final expressions for the belief update by removing the terms $I(\widehat{\theta}_j^{t-1}, \widehat{\theta}_j^t)$ and $I(\widehat{m}_j^{t-1}, \widehat{m}_j^t)$ and changing the scope of the first summations.

When $m_j$ in the interactive states is intentional:

$$\begin{aligned} b_i^t(is^t) &= \beta \sum_{is^{t-1}: \widehat{m}_j^{t-1} = \widehat{\theta}_j^t} b_i^{t-1}(is^{t-1}) \sum_{a_j^{t-1}} Pr(a_j^{t-1}|\theta_j^{t-1}) O_i(s^t, a^{t-1}, o_i^t) \\ &\times \sum_{o_j^t} \tau_{\theta_j^t}(b_j^{t-1}, a_j^{t-1}, o_j^t, b_j^t) O_j(s^t, a^{t-1}, o_j^t) T_i(s^{t-1}, a^{t-1}, s^t) \end{aligned} \tag{18}$$

Else, if it is subintentional:

$$\begin{aligned} b_i^t(is^t) &= \beta \sum_{is^{t-1}: \widehat{m}_j^{t-1} = \widehat{m}_j^t} b_i^{t-1}(is^{t-1}) \sum_{a_j^{t-1}} Pr(a_j^{t-1}|m_j^{t-1}) O_i(s^t, a^{t-1}, o_i^t) \\ &\times \sum_{o_j^t} \delta_K(\text{APPEND}(h_j^{t-1}, o_j^t), h_j^t) O_j(s^t, a^{t-1}, o_j^t) T_i(s^{t-1}, a^{t-1}, s^t) \end{aligned} \tag{19}$$

Since proposition 2 expresses the belief $b_i^t(is^t)$ in terms of parameters of the previous time step only, Proposition 1 holds as well. □

Before we present the proof of Theorem 1 we note that the Equation 7, which defines value iteration in I-POMDPs, can be rewritten in the following form, $U^n = HU^{n-1}$. Here, $H : B \rightarrow B$ is a *backup* operator, and is defined as,

$$HU^{n-1}(\theta_i) = \max_{a_i \in A_i} h(\theta_i, a_i, U^{n-1})$$

where $h : \Theta_i \times A_i \times B \rightarrow \mathbb{R}$ is,

$$h(\theta_i, a_i, U) = \sum_{is} b_i(is) ER_i(is, a_i) + \gamma \sum_{o \in \Omega_i} Pr(o_i|a_i, b_i) U(\langle SE_{\theta_i}(b_i, a_i, o_i), \hat{\theta}_i \rangle)$$

and where $B$ is the set of all bounded value functions $U$. Lemmas 1 and 2 establish important properties of the backup operator. Proof of Lemma 1 is given below, and proof of Lemma 2 follows thereafter.

*Proof of Lemma 1.* Select arbitrary value functions $V$ and $U$ such that $V(\theta_{i,l}) \leq U(\theta_{i,l}) \;\; \forall \theta_{i,l} \in \Theta_{i,l}$. Let $\theta_{i,l}$ be an arbitrary type of agent $i$.

$$
\begin{aligned}
HV(\theta_{i,l}) &= \max_{a_i \in A_i} \left\{ \sum_{is} b_i(is) ER_i(is, a_i) + \gamma \sum_{o \in \Omega_i} Pr(o_i|a_i, b_i) V(\langle SE_{\theta_{i,l}}(b_i, a_i, o_i), \hat{\theta}_i \rangle) \right\} \\
&= \sum_{is} b_i(is) ER_i(is, a_i^*) + \gamma \sum_{o \in \Omega_i} Pr(o_i|a_i^*, b_i) V(\langle SE_{\theta_{i,l}}(b_i, a_i^*, o_i), \hat{\theta}_i \rangle) \\
&\leq \sum_{is} b_i(is) ER_i(is, a_i^*) + \gamma \sum_{o \in \Omega_i} Pr(o_i|a_i^*, b_i) U(\langle SE_{\theta_{i,l}}(b_i, a_i^*, o_i), \hat{\theta}_i \rangle) \\
&\leq \max_{a_i \in A_i} \left\{ \sum_{is} b_i(is) ER_i(is, a_i) + \gamma \sum_{o \in \Omega_i} Pr(o_i|a_i, b_i) U(\langle SE_{\theta_{i,l}}(b_i, a_i, o_i), \hat{\theta}_i \rangle) \right\} \\
&= HU(\theta_{i,l})
\end{aligned}
$$

Since $\theta_{i,l}$ is arbitrary, $HV \leq HU$. □

*Proof of Lemma 2.* Assume two arbitrary well defined value functions $V$ and $U$ such that $V \leq U$. From Lemma 1 it follows that $HV \leq HU$. Let $\theta_{i,l}$ be an arbitrary type of agent $i$. Also, let $a_i^*$ be the action that optimizes $HU(\theta_{i,l})$.

$$
\begin{aligned}
0 &\leq HU(\theta_{i,l}) - HV(\theta_{i,l}) \\
&= \max_{a_i \in A_i} \left\{ sum_{is} b_i(is) ER_i(is, a_i) + \gamma \sum_{o \in \Omega_i} Pr(o_i|a_i, b_i) U(SE_{\theta_{i,l}}(b_i, a_i, o_i), \langle \hat{\theta}_i \rangle) \right\} - \\
&\quad \max_{a_i \in A_i} \left\{ \sum_{is} b_i(is) ER_i(is, a_i) + \gamma \sum_{o \in \Omega_i} Pr(o_i|a_i, b_i) V(SE_{\theta_{i,l}}(b_i, a_i, o_i), \langle \hat{\theta}_i \rangle) \right\} \\
&\leq \sum_{is} b_i(is) ER_i(is, a_i^*) + \gamma \sum_{o \in \Omega_i} Pr(o_i|a_i^*, b_i) U(SE_{\theta_{i,l}}(b_i, a_i^*, o_i), \langle \hat{\theta}_i \rangle) - \\
&\quad \sum_{is} b_i(is) ER_i(is, a_i^*) - \gamma \sum_{o \in \Omega_i} Pr(o_i|a_i^*, b_i) V(SE_{\theta_{i,l}}(b_i, a_i^*, o_i), \langle \hat{\theta}_i \rangle) \\
&= \gamma \sum_{o \in \Omega_i} Pr(o_i|a_i^*, b_i) U(SE_{\theta_{i,l}}(b_i, a_i^*, o_i), \langle \hat{\theta}_i \rangle) - \\
&\quad \gamma \sum_{o \in \Omega_i} Pr(o_i|a_i^*, b_i) V(SE_{\theta_{i,l}}(b_i, a_i^*, o_i), \langle \hat{\theta}_i \rangle) \\
&= \gamma \sum_{o \in \Omega_i} Pr(o_i|a_i^*, b_i) \left[ U(SE_{\theta_{i,l}}(b_i, a_i^*, o_i), \langle \hat{\theta}_i \rangle) - V(SE_{\theta_{i,l}}(b_i, a_i^*, o_i), \langle \hat{\theta}_i \rangle) \right\} \\
&\leq \gamma \sum_{o \in \Omega_i} Pr(o_i|a_i^*, b_i) ||U - V|| \\
&= \gamma ||U - V||
\end{aligned}
$$

As the supremum norm is symmetrical, a similar result can be derived for $HV(\theta_{i,l}) - HU(\theta_{i,l})$. Since $\theta_{i,l}$ is arbitrary, the Contraction property follows, i.e. $||HV - HU|| \leq ||V - U||$. □

Lemmas 1 and 2 provide the stepping stones for proving Theorem 1. Proof of Theorem 1 follows from a straightforward application of the Contraction Mapping Theorem. We state the Contraction Mapping Theorem (Stokey & Lucas, 1989) below:

**Theorem 3 (Contraction Mapping Theorem).** *If $(S, \rho)$ is a complete metric space and $T : S \to S$ is a contraction mapping with modulus $\gamma$, then*

1. *$T$ has exactly one fixed point $U^*$ in $S$, and*
2. *The sequence $\{U^n\}$ converges to $U^*$.*

Proof of Theorem 1 follows.

*Proof of Theorem 1.* The normed space $(B, ||\cdot||)$ is complete w.r.t the metric induced by the supremum norm. Lemma 2 establishes the contraction property of the backup operator, $H$. Using Theorem 3, and substituting $T$ with $H$, convergence of value iteration in I-POMDPs to a unique fixed point is established. □

We go on to the piecewise linearity and convexity (PWLC) property of the value function. We follow the outlines of the analogous proof for POMDPs in (Hausktecht, 1997; Smallwood & Sondik, 1973).

Let $\alpha : IS \rightarrow \mathbb{R}$ be a real-valued and bounded function. Let the space of such real-valued bounded functions be $B(IS)$. We will now define an inner product.

**Definition 5 (Inner product).** *Define the inner product, $\langle\cdot,\cdot\rangle : B(IS) \times \Delta(IS) \rightarrow \mathbb{R}$, by*

$$\langle \alpha, b_i \rangle = \sum_{is} b_i(is)\alpha(is)$$

The next lemma establishes the bilinearity of the inner product defined above.

**Lemma 3 (Bilinearity).** *For any $s, t \in \mathbb{R}$, $f, g \in B(IS)$, and $b, \lambda \in \Delta(IS)$ the following equalities hold:*

$$\langle sf + tg, b\rangle = s\langle f, b\rangle + t\langle g, b\rangle$$
$$\langle f, sb + t\lambda\rangle = s\langle f, b\rangle + t\langle f, \lambda\rangle$$

We are now ready to give the proof of Theorem 2. Theorem 4 restates Theorem 2 mathematically, and its proof follows thereafter.

**Theorem 4 (PWLC).** *The value function, $U^n$, in finitely nested I-POMDP is piece-wise linear and convex (PWLC). Mathematically,*

$$U^n(\theta_{i,l}) = \max_{\alpha^n} \sum_{is} b_i(is)\alpha^n(is) \qquad n = 1, 2, ...$$

*Proof of Theorem 4.* **Basis Step:** $n = 1$

From Bellman's Dynamic Programming equation,

$$U^1(\theta_i) = \max_{a_i} \sum_{is} b_i(is)ER(is, a_i) \tag{20}$$

where $ER_i(is, a_i) = \sum_{a_j} R(is, a_i, a_j)Pr(a_j|m_j)$. Here, $ER_i(\cdot)$ represents the expectation of $R$ w.r.t. agent $j$'s actions. Eq. 20 represents an inner product and using Lemma 3, the inner product is linear in $b_i$. By selecting the maximum of a set of linear vectors (hyperplanes), we obtain a PWLC horizon 1 value function.

**Inductive Hypothesis:** Suppose that $U^{n-1}(\theta_{i,l})$ is PWLC. Formally we have,

$$\begin{aligned} U^{n-1}(\theta_{i,l}) &= \max_{\alpha^{n-1}} \sum_{is} b_i(is)\alpha^{n-1}(is) \\ &= \max_{\dot\alpha^{n-1},\, \ddot\alpha^{n-1}} \left\{ \sum_{is:m_j \in IM_j} b_i(is)\dot\alpha^{n-1}(is) + \sum_{is:m_j\in SM_j} b_i(is)\ddot\alpha^{n-1}(is) \right\} \end{aligned} \tag{21}$$

**Inductive Proof:** To show that $U^n(\theta_{i,l})$ is PWLC.

$$U^n(\theta_{i,l}) = \max_{a_i^{t-1}} \left\{ \sum_{is^{t-1}} b_i^{t-1}(is^{t-1}) ER_i(is^{t-1}, a_i^{t-1}) + \gamma \sum_{o_i^t} Pr(o_i^t | a_i^{t-1}, b_i^{t-1}) U^{n-1}(\theta_{i,l}) \right\}$$

From the inductive hypothesis:

$$\begin{aligned} U^n(\theta_{i,l}) = \max_{a_i^{t-1}} \Big\{ & \textstyle\sum_{is^{t-1}} b_i^{t-1}(is^{t-1}) ER_i(is^{t-1}, a_i^{t-1}) \\ & + \gamma \textstyle\sum_{o_i^t} Pr(o_i^t | a_i^{t-1}, b_i^{t-1}) \max_{\alpha^{n-1} \in \Gamma^{n-1}} \sum_{is^t} b_i^t(is^t) \alpha^{n-1}(is^t) \Big\} \end{aligned}$$

Let $l(b_i^{t-1}, a_i^{t-1}, o_i^t)$ be the index of the alpha vector that maximizes the value at $b_i^t = SE(b_i^{t-1}, a_i^{t-1}, o_i^t)$. Then,

$$\begin{aligned} U^n(\theta_{i,l}) = \max_{a_i^{t-1}} \Big\{ & \textstyle\sum_{is^{t-1}} b_i^{t-1}(is^{t-1}) ER_i(is^{t-1}, a_i^{t-1}) \\ & + \gamma \textstyle\sum_{o_i^t} Pr(o_i^t | a_i^{t-1}, b_i^{t-1}) \sum_{is^t} b_i^t(is^t) \alpha^{n-1}_{l(b_i^{t-1}, a_i^{t-1}, o_i^t)} \Big\} \end{aligned}$$

From the second equation in the inductive hypothesis:

$$\begin{aligned} U^n(\theta_{i,l}) = \max_{a_i^{t-1}} \Big\{ & \textstyle\sum_{is^{t-1}} b_i^{t-1}(is^{t-1}) ER_i(is^{t-1}, a_i^{t-1}) + \gamma \sum_{o_i^t} Pr(o_i^t | a_i^{t-1}, b_i^{t-1}) \\ & \times \Big\{ \textstyle\sum_{is^t : m_j^t \in IM_j} b_i^t(is^t) \dot{\alpha}^{n-1}_{l(b_i^{t-1}, a_i^{t-1}, o_i^t)} + \sum_{is^t : m_j^t \in SM_j} b_i^t(is^t) \ddot{\alpha}^{n-1}_{l(b_i^{t-1}, a_i^{t-1}, o_i^t)} \Big\} \Big\} \end{aligned}$$

Substituting $b_i^t$ with the appropriate belief updates from Eqs. 17 and 17′ we get:

$$\begin{aligned} U^n(\theta_{i,l}) & = \max_{a_i^{t-1}} \Big\{ \textstyle\sum_{is^{t-1}} b_i^{t-1}(is^{t-1}) ER_i(is^{t-1}, a_i^{t-1}) + \gamma \sum_{o_i^t} Pr(o_i^t | a_i^{t-1}, b_i^{t-1}) \\ & \times \beta \Big[ \textstyle\sum_{is^t : m_j^t \in IM_j} \sum_{is^{t-1}} b_i^{t-1}(is^{t-1}) \Big\{ \sum_{a_j^{t-1}} Pr(a_j^{t-1} | \theta_j^{t-1}) \Big[ O_i(s^t, a^{t-1}, o_i^t) \\ & \times \textstyle\sum_{o_j^t} O_j^t(s^t, a^{t-1}, o_j^t) \Big\{ \tau_{\theta_j^t}(b_j^{t-1}, a_j^{t-1}, o_j^t, b_j^t) I(\widehat{\theta}_j^{t-1}, \widehat{\theta}_j^t) T_i(s^{t-1}, a^{t-1}, s^t) \Big\} \Big] \Big\} \\ & \times \dot{\alpha}^{n-1}_{l(b_i^{t-1}, a_i^{t-1}, o_i^t)}(is^t) \\ & + \textstyle\sum_{is^t : m_j^t \in SM_j} \sum_{is^{t-1}} b_i^{t-1}(is^{t-1}) \Big\{ \sum_{a_j^{t-1}} Pr(a_j^{t-1} | m_j^{t-1}) \Big[ O_i(s^t, a^{t-1}, o_i^t) \\ & \times \textstyle\sum_{o_j^t} O_j^t(s^t, a^{t-1}, o_j^t) \Big\{ \delta_K(\text{APPEND}(h_j^{t-1}, o_j^t) - h_j^t) I(\widehat{m}_j^{t-1}, \widehat{m}_j^t) T_i(s^{t-1}, a^{t-1}, s^t) \Big\} \Big] \Big\} \\ & \times \ddot{\alpha}^{n-1}_{l(b_i^{t-1}, a_i^{t-1}, o_i^t)}(is^t) \Big] \Big\} \end{aligned}$$

and further

$$
\begin{aligned}
U^n(\theta_{i,l}) \;&= \max_{a_i^{t-1}} \Bigg\{ \sum_{is^{t-1}} b_i^{t-1}(is^{t-1}) ER_i(is^{t-1}, a_i^{t-1}) + \gamma \sum_{o_i^t} \Bigg[ \sum_{is^t: m_j^t \in IM_j} \\
&\times \sum_{is^{t-1}} b_i^{t-1}(is^{t-1}) \Bigg\{ \sum_{a_j^{t-1}} Pr(a_j^{t-1}|\theta_j^{t-1}) \Bigg[ O_i(s^t, a^{t-1}, o_i^t) \\
&\times \sum_{o_j^t} O_j^t(s^t, a^{t-1}, o_j^t) \Bigg\{ \tau_{\theta_j^t}(b_j^{t-1}, a_j^{t-1}, o_j^t, b_j^t) I(\widehat{\theta}_j^{t-1}, \widehat{\theta}_j^t) T_i(s^{t-1}, a^{t-1}, s^t) \Bigg\} \Bigg] \Bigg\} \\
&\times \dot{\alpha}^{n-1}_{l(b_i^{t-1}, a_i^{t-1}, o_i^t)}(is^t) \\
&+ \sum_{is^t: m_j^t \in SM_j} \sum_{is^{t-1}} b_i^{t-1}(is^{t-1}) \Bigg\{ \sum_{a_j^{t-1}} Pr(a_j^{t-1}|m_j^{t-1}) \Bigg[ O_i(s^t, a^{t-1}, o_i^t) \\
&\times \sum_{o_j^t} O_j^t(s^t, a^{t-1}, o_j^t) \Bigg\{ \delta_K(\text{APPEND}(h_j^{t-1}, o_j^t) - h_j^t) I(\widehat{m}_j^{t-1}, \widehat{m}_j^t) T_i(s^{t-1}, a^{t-1}, s^t) \Bigg\} \Bigg] \Bigg\} \\
&\times \ddot{\alpha}^{n-1}_{l(b_i^{t-1}, a_i^{t-1}, o_i^t)}(is^t) \Bigg] \Bigg\}
\end{aligned}
$$

Rearranging the terms of the equation:

$$
\begin{aligned}
U^n(\theta_{i,l}) \;&= \max_{a_i^{t-1}} \Bigg\{ \sum_{is^{t-1}: m_j^{t-1} \in IM_j} b_i^{t-1}(is^{t-1}) \Bigg\{ ER_i(is^{t-1}, a_i^{t-1}) + \gamma \sum_{o_i^t} \sum_{is^t: m_j^t \in IM_j} \\
&\times \Bigg\{ \sum_{a_j^{t-1}} Pr(a_j^{t-1}|\theta_j^{t-1}) \Bigg[ O_i(s^t, a^{t-1}, o_i^t) \sum_{o_j^t} O_j^t(s^t, a^{t-1}, o_j^t) \\
&\times \Bigg\{ \tau_{\theta_j^t}(b_j^{t-1}, a_j^{t-1}, o_j^t, b_j^t) I(\widehat{\theta}_j^{t-1}, \widehat{\theta}_j^t) T_i(s^{t-1}, a^{t-1}, s^t) \Bigg\} \Bigg] \Bigg\} \dot{\alpha}^{n-1}_{l(b_i^{t-1}, a_i^{t-1}, o_i^t)}(is^t) \Bigg\} \\
&+ \sum_{is^{t-1}: m_j^{t-1} \in SM_j} b_i^{t-1}(is^{t-1}) \Bigg\{ ER_i(is^{t-1}, a_i^{t-1}) + \gamma \sum_{o_i^t} \sum_{is^t: m_j^t \in SM_j} \sum_{o_i^t} \\
&\times \Bigg\{ \sum_{a_j^{t-1}} Pr(a_j^{t-1}|m_j^{t-1}) \Bigg[ O_i(s^t, a^{t-1}, o_i^t) \sum_{o_j^t} O_j^t(s^t, a^{t-1}, o_j^t) \\
&\times \Bigg\{ \delta_K(\text{APPEND}(h_j^{t-1}, o_j^t) - h_j^t) I(\widehat{m}_j^{t-1}, \widehat{m}_j^t) T_i(s^{t-1}, a^{t-1}, s^t) \Bigg\} \Bigg] \Bigg\} \ddot{\alpha}^{n-1}_{l(b_i^{t-1}, a_i^{t-1}, o_i^t)}(is^t) \Bigg\} \Bigg\} \\
&= \max_{a_i^{t-1}} \Bigg\{ \sum_{is^{t-1}: m_j^{t-1} \in IM_j} b_i^{t-1}(is^{t-1}) \dot{\alpha}^n_{a_i}(is^{t-1}) \\
&+ \sum_{is^{t-1}: m_j^{t-1} \in SM_j} b_i^{t-1}(is^{t-1}) \ddot{\alpha}^n_{a_i}(is^{t-1}) \Bigg\}
\end{aligned}
$$

Therefore,

$$
\begin{aligned}
U^n(\theta_{i,l}) \;&= \max_{\dot{\alpha}^n,\, \ddot{\alpha}^n} \Bigg\{ \sum_{is^{t-1}: m_j^{t-1} \in IM_j} b_i^{t-1}(is^{t-1}) \dot{\alpha}^n(is^{t-1}) \\
&+ \sum_{is^{t-1}: m_j^{t-1} \in SM_j} b_i^{t-1}(is^{t-1}) \ddot{\alpha}^n(is^{t-1}) \Bigg\} \\
&= \max_{\alpha^n} \sum_{is^{t-1}} b_i^{t-1}(is^{t-1}) \alpha^n(is^{t-1}) = \max_{\alpha^n} \langle b_i^{t-1}, \alpha^n \rangle
\end{aligned} \qquad (22)
$$

where, if $m_j^{t-1}$ in $is^{t-1}$ is intentional then $\alpha^n = \dot{\alpha}^n$:

$$\begin{aligned}\dot{\alpha}^n(is^{t-1}) &= ER_i(is^{t-1}, a_i^{t-1}) + \gamma \sum_{o_i^t} \sum_{is^t : m_j^t \in IM_j} \Bigg\{ \sum_{a_j^{t-1}} Pr(a_j^{t-1}|\theta_j^{t-1}) \Bigg[ O_i(is^t, a^{t-1}, o_i^t) \\ &\times \sum_{o_j^t} O_j^t(is_j^t, a^{t-1}, o_j^t) \Bigg\{ \tau_{\theta_j^t}(b_j^{t-1}, a_j^{t-1}, o_j^t, b_j^t) I(\widehat{\theta}_j^{t-1}, \widehat{\theta}_j^t) T_i(s^{t-1}, a^{t-1}, s^t) \Bigg\} \Bigg] \Bigg\} \\ &\times \alpha^{n-1}_{l(b_i^{t-1}, a_i^{t-1}, o_i^t)}(is^t)\end{aligned}$$

and, if $m_j^{t-1}$ is subintentional then $\alpha^n = \ddot{\alpha}^n$:

$$\begin{aligned}\ddot{\alpha}^n(is^{t-1}) &= ER_i(is^{t-1}, a_i^{t-1}) + \gamma \sum_{o_i^t} \sum_{is^t : m_j^t \in SM_j} \Bigg\{ \sum_{a_j^{t-1}} Pr(a_j^{t-1}|\theta_j^{t-1}) \Bigg[ O_i(is^t, a^{t-1}, o_i^t) \\ &\times \sum_{o_j^t} O_j^t(is_j^t, a^{t-1}, o_j^t) \Bigg\{ \delta_K(\text{APPEND}(h_j^{t-1}, o_j^t) - h_j^t) I(\widehat{m}_j^{t-1}, \widehat{m}_j^t) T_i(s^{t-1}, a^{t-1}, s^t) \Bigg\} \Bigg] \Bigg\} \\ &\times \alpha^{n-1}_{l(b_i^{t-1}, a_i^{t-1}, o_i^t)}(is^t)\end{aligned}$$

Eq. 22 is an **inner product** and using Lemma 3, the value function is linear in $b_i^{t-1}$. Furthermore, maximizing over a set of linear vectors (hyperplanes) produces a piecewise linear and convex value function. □